\pdfoutput=1
\documentclass[11pt]{article}
\usepackage{algorithm}
\usepackage{algpseudocode}
\usepackage{amsmath}
\usepackage{booktabs}
\usepackage{multirow}
\usepackage{tcolorbox} 
\usepackage{times}
\usepackage{latexsym}
\usepackage{subcaption}
\usepackage[T1]{fontenc}
\usepackage[utf8]{inputenc}
\usepackage{microtype}
\usepackage{tabularx}
\usepackage{inconsolata}
\usepackage{graphicx}
\usepackage[final]{acl}
\usepackage{diagbox}
\usepackage{array}
\usepackage{caption}
\usepackage{multicol}
\usepackage{xcolor}

\title{Uncovering the Bigger Picture: Comprehensive Event Understanding Via Diverse News Retrieval}

\author{Yixuan Tang$^{1}$ \qquad Yuanyuan Shi$^{2}$\thanks{Work done while exchanging at NUS.} \qquad Yiqun Sun$^{1}$ \qquad  Anthony K.H. Tung$^{1}$\\ \\
  $^{1}$ National University of Singapore \qquad $^{2}$ Shanghai Jiao Tong University\\
  \texttt{\{yixuan, sunyq, atung\}@comp.nus.edu.sg \qquad syysyysyy1010@sjtu.edu.cn} \\ }

\begin{document}
\maketitle
\begin{abstract}

Access to diverse perspectives is essential for understanding real-world events, yet most news retrieval systems prioritize textual relevance, leading to redundant results and limited viewpoint exposure. We propose \textbf{NEWSCOPE}, a two-stage framework for diverse news retrieval that enhances event coverage by explicitly modeling semantic variation at the sentence level. The first stage retrieves topically relevant content using dense retrieval, while the second stage applies sentence-level clustering and diversity-aware re-ranking to surface complementary information. To evaluate retrieval diversity, we introduce three interpretable metrics, namely \textit{Average Pairwise Distance}, \textit{Positive Cluster Coverage}, and \textit{Information Density Ratio}, and construct two paragraph-level benchmarks: \textbf{LocalNews} and \textbf{DSGlobal}. Experiments show that \textbf{NEWSCOPE} consistently outperforms strong baselines, achieving significantly higher diversity without compromising relevance. Our results demonstrate the effectiveness of fine-grained, interpretable modeling in mitigating redundancy and promoting comprehensive event understanding. The data and code are available at \url{https://github.com/tangyixuan/NEWSCOPE}.

\end{abstract}

\section{Introduction}

\begin{figure}[t]
\centering
\includegraphics[width=0.95 \columnwidth]{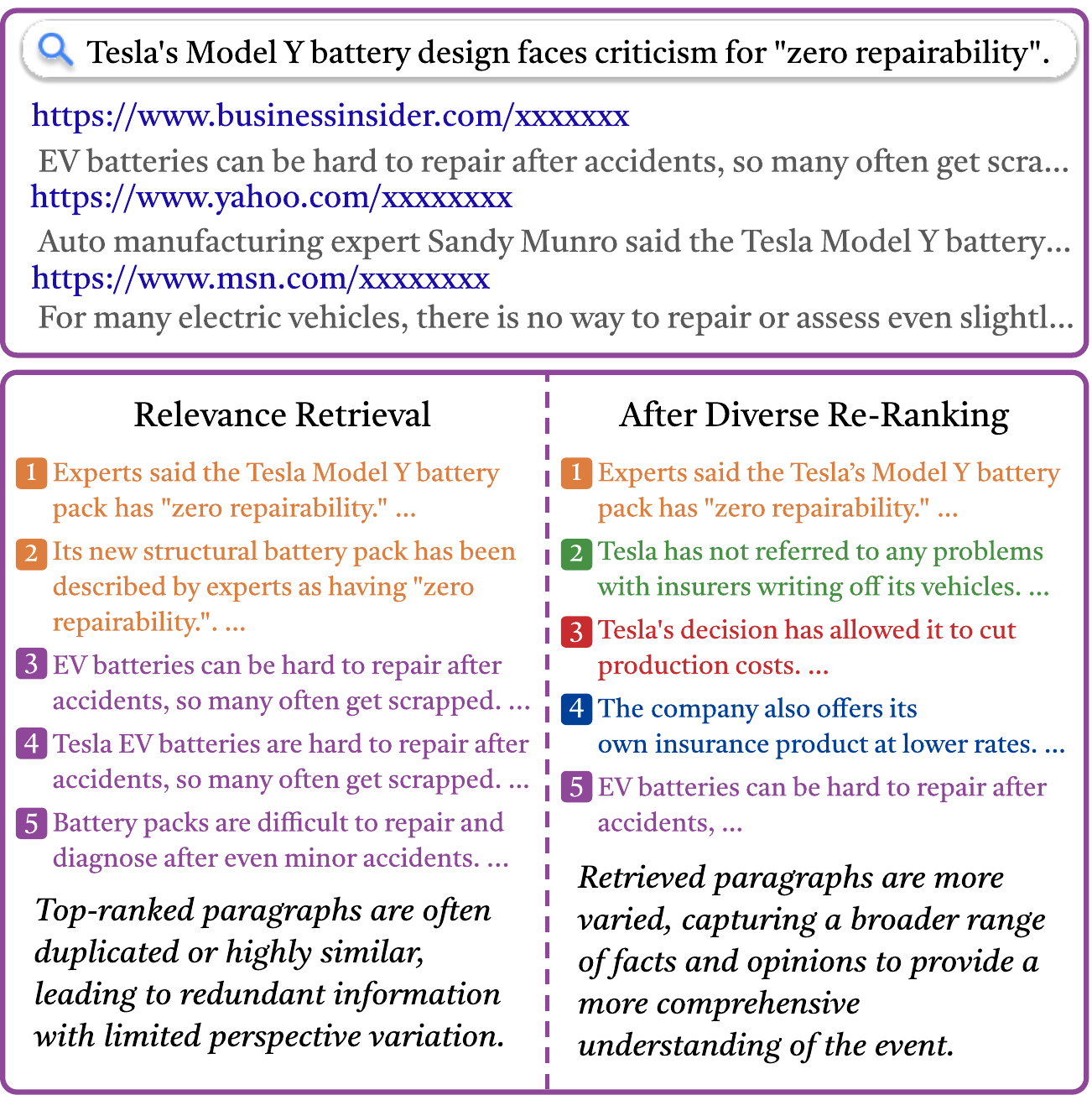}
\caption{Comparison of dense and diverse retrieval results. Dense retrieval returns similar and redundant paragraphs, while diverse retrieval captures relevant yet distinct paragraphs, improving coverage of different perspectives. Same color indicates semantic similarity.}

\label{fig:eg}
\end{figure}

The ability to access to diverse perspectives is fundamental for understanding complex real-world events. However, most existing systems prioritize textual relevance, often producing results that are not only biased toward the query but also highly similar to one another~\cite{kulshrestha2019search}. This combination of bias and redundancy limits users' exposure to alternative viewpoints~\cite{spinde2022we}. As algorithm-driven news consumption becomes mainstream, relevance-focused retrieval risks amplifying echo chambers. This raises a critical question: how can retrieval systems surface distinct, complementary perspectives rather than reiterate the same narrative across sources?

Traditional text retrieval models, including BM25~\cite{robertson1995okapi} and dense retrievers~\cite{karpukhin-etal-2020-dense,thakur2beir}, prioritize textual relevance, often favoring lexically similar documents and repeating prominent entities~\cite{fayyaz2025collapse}. While effective for topical matching, they yield semantically overlapping results. Re-ranking methods like MMR~\cite{carbonell1998use} introduce diversity penalties, but operate at a coarse, article-level embedding representation and fail to capture fine-grained differences in perspectives, framing, and coverage. As shown in Figure~\ref{fig:eg}, relevance retrieval methods often produce near-duplicate content, limiting comprehensive understanding of events from complementary viewpoints.

Diversified retrieval has been explored in recommendation systems~\cite{DBLP:conf/wsdm/YuMPW14,DBLP:conf/nips/ChenZZ18,DBLP:conf/wsdm/ZhangW023}, where diversity is defined via user behavior or product categories. However, such domain-specific notions do not directly transfer to news retrieval, where diversity involves distinct factual aspects and viewpoints. Existing methods~\cite{DBLP:conf/sigir/XiaXLGC16,DBLP:conf/sigir/JiangWDZNY17} are typically item-level and supervised, limiting interpretability and scalability. These limitations give rise to two core research questions:

\vspace{0.5em}
\noindent\textbf{RQ1}: \textit{How can we systematically define and measure diversity in news retrieval?}\\
\textbf{RQ2}: \textit{How can we balance high relevance with comprehensive coverage of distinct perspectives?}
\vspace{0.5em}

To this end, we introduce the task of \textbf{event-centric diverse news retrieval}, which aims to retrieve relevant documents that collectively represent multiple comprehensive perspectives on a target event. We treat sentences as the atomic units of semantic meaning and model diversity at the sentence level to capture fine-grained distinctions. We operate retrieval at the paragraph level to ensure output coherence. To evaluate retrieval diversity, we propose three interpretable metrics: \textit{Average Pairwise Distance} (D), \textit{Positive Cluster Coverage} (C), and \textit{Information Density Ratio} (I), which capture perspective coverage and semantic richness beyond standard precision and recall.

Motivated by these challenges, we propose \textbf{NEWSCOPE} (\textit{\textbf{NEW}s Understanding via \textbf{S}entence \textbf{C}lustering and c\textbf{O}m\textbf{P}rehensive r\textbf{E}trieval)}, a two-stage diversified retrieval framework. The first stage efficiently retrieves candidate documents based on relevance scoring. The second stage applies sentence-level semantic clustering and diversity-aware re-ranking to construct a set of paragraphs that are both relevant and perspectively diverse. To support the task, we construct two paragraph-annotated benchmarks: \textbf{LocalNews}, focused on regional reporting, and \textbf{DSGlobal}, an adaptation of the DiverseSumm summarization corpus~\cite{DBLP:conf/naacl/HuangLFCJXW24} focused on global news for retrieval evaluation. Experiments show that \textbf{NEWSCOPE} consistently outperforms strong baselines, achieving significantly higher diversity without compromising relevance, while maintaining practical efficiency.

Our main contributions are as follows:
\begin{itemize}
    \item We formalize the task of \textbf{event-centric diverse news retrieval} and propose three novel \textbf{interpretable metrics} to evaluate the diversity of retrieved content, capturing fine-graind information coverage.
    \item We propose \textbf{NEWSCOPE}, a two-stage retrieval framework that enhance dense relevance retrieval with sentence-level clustering and diversity-aware re-ranking to promote comprehensive perspective coverage.
    \item We curate two paragraph-level benchmarks, \textbf{LocalNews} and \textbf{DSGlobal} to support evaluation. We show that \textbf{NEWSCOPE} consistently outperforms strong baselines on both datasets.
\end{itemize}

Our work highlights the importance of fine-grained and interpretable diversity modeling for news retrieval, aiming to break echo chambers and support more comprehensive, unbiased event understanding. While we focus on the news domain where divergent perspectives and conflicting narratives are common, the proposed framework is domain-agnostic and can be extended to other contexts where diverse information is valuable.

\section{Related Work}

\begin{figure*}[th]
\centering
  \includegraphics[width=0.95\linewidth]{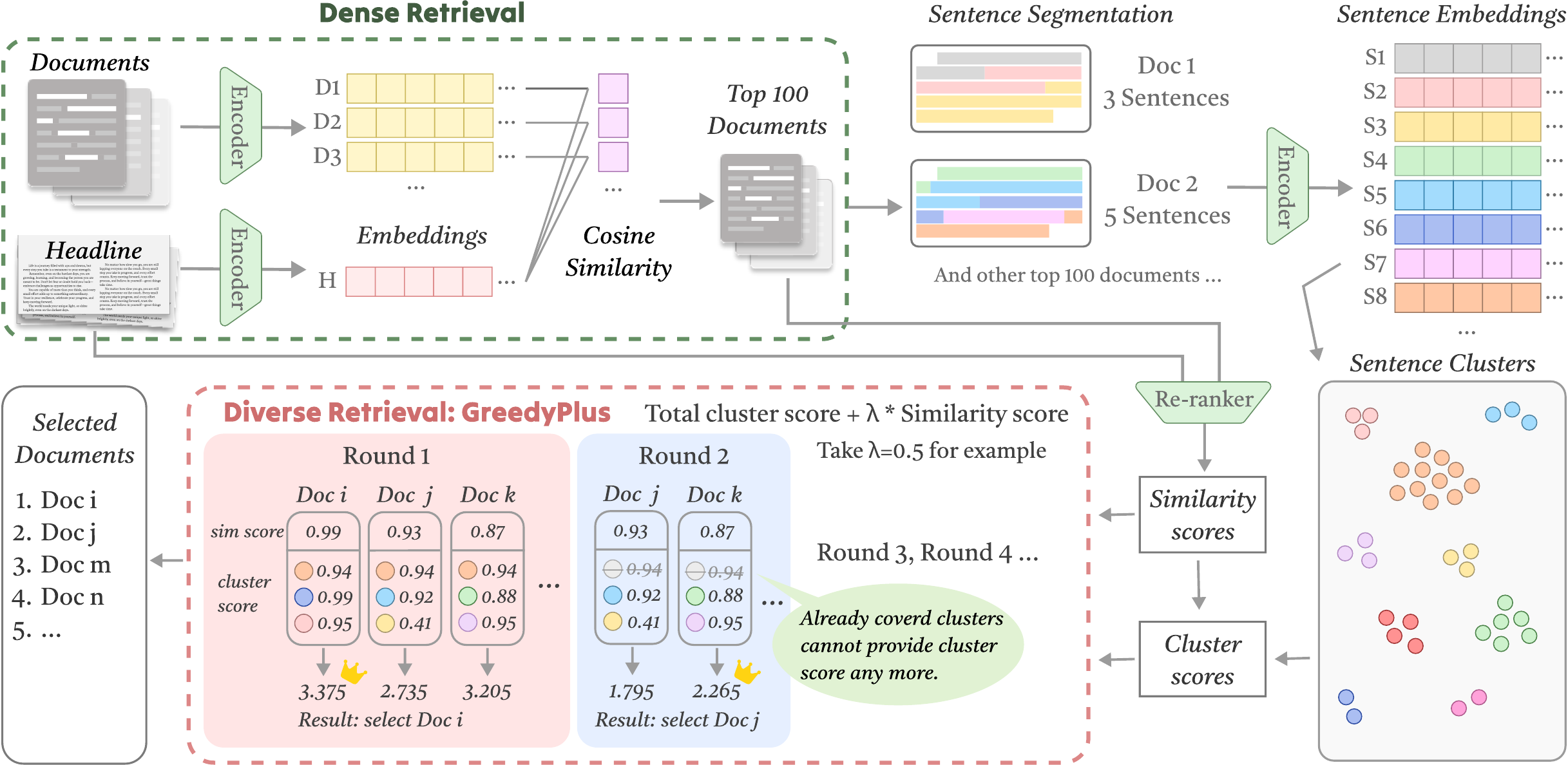} 
\caption{Overview of the \textbf{NEWSCOPE} framework. The two-stage pipeline combines efficient dense retrieval with sentence-level diversity-aware re-ranking to ensure both relevance and comprehensive perspective coverage.}
    \label{fig:overview}
\end{figure*}

\subsection{Comprehensive News Coverage}

Comprehensive news understanding requires aggregating perspectives from multiple sources. Traditional fact-checking and misinformation detection work has focused on claim-level veracity~\citep{DBLP:conf/naacl/ThorneVCM18}, often ignoring source bias. Recent studies warn of selective exposure to biased narratives~\citep{DBLP:conf/aaai/EstornellDV20,DBLP:conf/kdd/FungHNJ22,DBLP:journals/corr/abs-2308-07973,DBLP:journals/eswa/RodrigoGinesCP24,DBLP:journals/corr/abs-2312-16490, DBLP:conf/emnlp/TRACER}. Retrieval-augmented generation (RAG) pipelines have been adopted for evidence-backed verification~\citep{DBLP:journals/corr/abs-2305-16128,DBLP:journals/corr/abs-2408-12060,DBLP:journals/corr/abs-2410-04657,DBLP:journals/corr/abs-2412-15189}, yet these focus on factual correctness over viewpoint diversity.

Recent work on social media analysis has shown that capturing a diverse range of viewpoints is essential for understanding the full spectrum of events \citep{DBLP:journals/corr/abs-2408-04463,DBLP:journals/corr/abs-2408-14974,DBLP:journals/corr/abs-2409-08829}. Some studies focus on aligning sentence-level information across different perspectives or generating summaries of news events but do not incorporate retrieval mechanisms. The methods often assume access to balanced inputs for each event and lack explicit retrieval components. \citep{DBLP:journals/corr/abs-2401-05650,DBLP:conf/acl/ChenTT24,DBLP:conf/naacl/ZhangZ0FLKLXZRM24,DBLP:conf/naacl/HuangLFCJXW24,DBLP:conf/acl/FabbriLSLR19}, which motivate the need for retrieval strategies that ensure balanced and diverse event representations.

\subsection{Diverse Retrieval}

Traditional retrieval models such as BM25~\cite{robertson1995okapi} and dense retrievers~\cite{thakur2beir, muennighoff-etal-2023-mteb} prioritize textual similarity, often surfacing redundant content that repeats dominant narratives. Maximal Marginal Relevance (MMR)~\citep{carbonell1998use} introduces a relevance–diversity trade-off and has been adapted for summarization and dialogue tracking~\citep{DBLP:conf/acl/FabbriLSLR19,DBLP:conf/acl/KingF23}, but it lacks explicit modeling of fine-grained semantic coverage and interpretability.

Diversity-oriented retrieval has been widely explored in other domains~\cite{DBLP:conf/wsdm/YuMPW14}. DPP-based models~\citep{DBLP:conf/nips/ChenZZ18} and disentangled representations~\citep{DBLP:conf/wsdm/ZhangW023} are used in recommendation systems to reduce redundancy across items. They rely on user profiles such as purchase history and are evaluated on e-commerce, music, or video platforms, making them less applicable to news retrieval. Other methods~\citep{DBLP:conf/sigir/XiaXLGC16, DBLP:conf/sigir/JiangWDZNY17} rely on supervised models and subtopic attention mechanisms, requiring large amount of labeled data for training. In parallel, text enrichment frameworks such as QALink~\cite{DBLP:conf/cikm/TangHLTWYZ17} and QALinkPlus~\cite{DBLP:journals/debu/SunTT23} demonstrate that diverse selection of QA content can supplement missing background knowledge, highlighting the value of fine-grained, interpretable diversified information enrichment.

Within news retrieval, D$k$MIPS~\citep{DBLP:journals/corr/abs-2402-13858} promotes stance diversity by penalizing similarity in dense embeddings. It struggles with fine-grained diversity control and lacks semantic transparency. PerspectroScope~\citep{DBLP:conf/acl/ChenKCR19} targets claim-centric stance discovery by identifying supporting and opposing views. In contrast, our setting is event-centric and stance-agnostic. We introduce an unsupervised, interpretable framework for news retrieval. It jointly optimizes relevance and sentence-level diversity without requiring domain-specific training or annotated stances, making it more generalizable. 

\section{Problem Formulation}

Given a target event $E$ and a large news corpus $\mathcal{C}$, the goal of \textbf{event-centric diverse news retrieval} is to retrieve a set of $k$ paragraphs $\mathcal{R} = \{r_1, r_2, \dots, r_k\}$ from $\mathcal{C}$ such that:

\vspace{0.2em}
\noindent \textbf{Relevance:} Each paragraph $r_i \in \mathcal{R}$ is topically relevant to the event $E$.

\noindent \textbf{Diversity:} The set $\mathcal{R}$ collectively covers semantically distinct aspects or perspectives of $E$, minimizing redundancy and maximizing subtopic coverage.

We treat paragraphs as the retrieval unit and model diversity at the sentence level to capture fine-grained variation between results. This enables the retrieval of complementary viewpoints that enhance the comprehensiveness of information presented to the user.

\section{The NEWSCOPE Framework}

We propose \textbf{NEWSCOPE}, a two-stage framework designed to retrieve relevant news content that collectively reflects a diverse set of perspectives on a given event. The first stage performs relevance-based retrieval using dense embeddings to collect candidate paragraphs. The second stage performs diversity-aware reranking by modeling sentence-level semantic variation and selecting content that maximizes both coverage and informativeness. Figure~\ref{fig:overview} provides an overview of the framework.

\subsection{Stage I: Relevance-Based Retrieval}

Given an event headline, we first retrieve a candidate set of paragraphs using a dense embedding-based retriever. Specifically, both the headline and candidate paragraphs are encoded using the \texttt{bilingual-embedding-large} model. Paragraphs are ranked by their cosine similarity to the headline in the embedding space. This step ensures that only highly relevant content is forwarded to the reranking stage, filtering out unrelated or low-quality paragraphs.

\subsection{Stage II: Diversity-Aware Reranking}
To capture diverse viewpoints, we rerank the candidate paragraphs by explicitly modeling sentence-level semantic diversity. This stage comprises three steps: (1) sentence clustering to identify semantically similar content, (2) greedy selection to cover distinct clusters, and (3) optional cluster weighting to jointly optimize diversity and relevance.

\subsubsection{Sentence Clustering}

We segment each paragraph in the candidate set into individual sentences, treating sentences as atomic meaning units. Each sentence is encoded as a dense vector. To uncover semantic similarity, we apply the OPTICS clustering algorithm \cite{ankerst1999optics}, which groups semantically related sentences based on density without requiring a predefined number of clusters. The resulting clusters capture distinct informational aspects of the event. Appendix~\ref{app:5cluster} presents the first five clusters from an example case, illustrating that sentence clusters are good proxies for event aspects.

\subsubsection{Greedy Cluster Selection (GreedySCS)}

To promote diversity, we employ a greedy selection strategy that iteratively selects paragraphs covering the most novel sentence clusters. Let \( P \) be the set of candidate paragraphs and \( C \) the set of sentence clusters. We initialize the selected paragraph set \( S \leftarrow \emptyset \) and the uncovered cluster set \( U \leftarrow C \). At each step, we score each paragraph by the number of uncovered clusters it contains and greedily select the one with the highest score. Formally, 
\begin{equation*}
\text{Score}(p) = |U \cap \text{Clusters}(p)|,
\end{equation*}
where $\text{Clusters}(p)$ denotes the cluster indices that the contains at least one sentence from the paragraph, i.e. $\text{Clusters}(p) = \{ C_{s_i} \mid s_i \in p \}$. The selected paragraph is added to \( S \), and its corresponding clusters are marked as covered. The process continues until a stopping condition is met: either a predefined number of paragraphs is selected or sufficient cluster coverage is achieved. This process is outlined in Algorithm~\ref{alg:greedy}. This method, referred to as \textbf{GreedySCS}, mitigates information redundancy and promotes a broad representation of perspectives by prioritizing novel clusters.

\begin{algorithm}[t]
\small
\caption{Greedy Cluster Selection}
\label{alg:greedy}
\begin{algorithmic}[1]
\Require $P$: Set of candidate paragraphs; $C$: Sentence clusters
\State Initialize $S \leftarrow \emptyset$, $U \leftarrow C$
\While{$|S| > k$ \textbf{or} $|U| > \text{coverage\_threshold}$}
    \For{$p \in P$}
        \State $\text{Score}(p) \leftarrow |U \cap \text{Clusters}(p)|$
    \EndFor
    \State $p^* \leftarrow \arg\max_p \text{Score}(p)$
    \State $S \leftarrow S \cup \{p^*\}$, $U \leftarrow U \setminus \text{Clusters}(p^*)$
\EndWhile
\State \Return $S$
\end{algorithmic}
\end{algorithm}

\subsubsection{Cluster-Based Weighting (GreedyPlus)}

To further refine paragraph selection, we enhance GreedySCS with a soft weighting mechanism that jointly models relevance and diversity. Specifically, we assign an importance score to each sentence cluster based on its overall relevance to the query.

\paragraph{Cluster Score} For each sentence cluster \( c \), we compute a relevance-based weight by averaging the similarity between the headline \( h \) and all paragraphs containing sentences from that cluster:
\begin{equation*}
\text{ClusterScore}(c) = \frac{1}{|c|} \sum_{s_i \in c} \text{Sim}(h, p_{s_i}),
\end{equation*}
where \( s_i \) is a sentence in cluster \( c \), \( p_{s_i} \) is the paragraph containing \( s_i \), and \( \text{Sim}(h, p_{s_i}) \) is the similarity score computed using the \texttt{bge-reranker-large} model. \( |c| \) is the number of sentences contained in this cluster.

The cluster score dynamically weights the importance and relevance of sentence clusters with respect to the query headline, serving as a fine-grained, soft noise filter that downweights low-quality or off-topic clusters present in the candidate pool from Stage I, such as incidental content or promotional text.

\paragraph{Paragraph Score} Each paragraph \( p \) is then scored by combining its relevance to the headline and its contribution to covering high-quality, previously unselected clusters \( U \):

\begin{align*}
& \text{Score}^+(p) = \\
& \underbrace{\sum_{c \in \text{clusters}(p)\cap U} \text{ClusterScore}(c)}_{\text{Diversity Term}} + \lambda \cdot \underbrace{\text{Sim}(h, p)}_{\text{Relevance Term}}
\end{align*}

The diversity term encourages the selection of paragraphs that introduce novel content by covering high-quality clusters that have not yet been selected, while the relevance term ensures alignment with the user query. The parameter \( \lambda \) controls the trade-off between diversity and relevance.

This enhanced version, denoted as \textbf{GreedyPlus}, leverages soft weighting to prioritize both informative and relevant content, resulting in greater alignment with the user’s query while maintaining viewpoint diversity.

\paragraph{Efficiency}
Since Stage II operates on a small set of top-ranked candidate paragraphs (e.g., 100) from Stage I for fine-grained re-ranking, the additional cost of diversity-aware re-ranking remains relatively low. Our pipeline runs efficiently, with an average runtime of approximately 1.2 seconds per event (see Appendix~\ref{sec:appendix_efficiency} for details).

\section{Data Construction}

To evaluate diverse news retrieval, we construct two benchmark datasets: \textbf{LocalNews}, a new corpus of paragraph-annotated local news events, and \textbf{DSGlobal}, an adaptation of the DiverseSumm dataset \citep{DBLP:conf/naacl/HuangLFCJXW24} for retrieval. These datasets cover both local and global contexts and enable robust, fine-grained evaluation of diversity-aware retrieval systems.

\subsection{Local News Benchmark (LocalNews)}

\begin{table}[t]
\small
\centering
\begin{tabular}{c c c}
\toprule
\textbf{Properties} & \textbf{LocalNews} & \textbf{DSGlobal} \\ \midrule
\textbf{\# Events} & 103 & 147 \\
\textbf{\# Paragraphs} & 5,296 & 7,532 \\
\textbf{Avg. Sentence Count} & 7.1 & 7.5 \\
\textbf{Avg. Word Count} & 124.9 & 123 \\
\bottomrule
\end{tabular}
\caption{Data Statistics for LocalNews and DSGlobal.}
\label{tab:stats}
\end{table}

\textbf{LocalNews} is built using Google News’ “Full Coverage” feature to collect multi-source reports on the same event. We apply event de-duplication and segment articles into paragraphs capped at 512 tokens, preserving semantic boundaries. The final set includes 5,296 paragraphs averaging 7.1 sentences and 124.9 words each.

Each paragraph was labeled for relevance against a one-sentence abstractive event summary generated by \texttt{GPT-4o-mini}, which served as the query for simulating user retrieval intent. Summaries were iteratively refined up to five times if their associated relevance labels had low positive rates. Paragraphs were annotated in three rounds with majority voting (98.7\% agreement), and those from outside the event’s full coverage were auto-labeled as irrelevant. A human review of 100 samples confirmed the accuracy of the relevance annotation, with a 99\% acceptance rate.

\begin{table*}[t]
\centering
\small
\begin{tabular}{@{}c cccccc cccccc@{}}
\toprule
\multirow{3}{*}{\textbf{Model}} 
& \multicolumn{3}{c}{\textbf{Relevancy}} & \multicolumn{3}{c}{\textbf{Diversity}} 
& \multicolumn{3}{c}{\textbf{Relevancy}} & \multicolumn{3}{c}{\textbf{Diversity}} \\
\cmidrule(lr){2-4} \cmidrule(lr){5-7} \cmidrule(lr){8-10} \cmidrule(lr){11-13}
 & \textbf{P} & \textbf{R} & \textbf{F1} & \textbf{D} & \textbf{I} & \textbf{C} & \textbf{P} & \textbf{R} & \textbf{F1} & \textbf{D} & \textbf{I} & \textbf{C} \\ 
 \cmidrule(lr){2-7}\cmidrule(lr){8-13}
 & \multicolumn{6}{c}{\textbf{top 5}} & \multicolumn{6}{c}{\textbf{top 10}} \\ 
 \midrule
\textbf{BM25} & 95.9 & 16.9 & 28.8 & 20.8 & 47.3 & 35.9 & 91.4 & 31.3 & 46.7 & 26.5 & 36.9 & 51.2 \\
\textbf{DenseRetr} & \textbf{97.5} & \textbf{17.2} & \textbf{29.3} & 17.0 & 45.2 & 34.2 & \textbf{95.0} & \textbf{32.9} & \textbf{48.9} & 23.0 & 36.5 & 52.8 \\
\textbf{MMR} & 93.0 & 16.6 & 28.1 & 29.8 & 54.3 & 41.3 & 92.8 & 32.4 & 48.1 & 29.0 & 38.8 & 55.8 \\
\textbf{D$k$MIPS} & 83.7 & 14.6 & 24.9 & \textbf{33.4} & 50.7 & 32.7 & 83.6 & 28.9 & 43.0 & \textbf{34.6} & 39.1 & 50.2 \\ \midrule
\textbf{NEWSCOPE (GreedySCS)} & 95.3 & 16.8 & 28.6 & 24.2 & 64.0 & 52.0 & 93.7 & 32.6 & 48.3 & 26.3 & 44.6 & 62.0 \\
\textbf{NEWSCOPE (GreedyPlus)} & 92.4 & 16.2 & 27.5 & 29.8 & \textbf{73.6} & \textbf{62.2} & 90.3 & 31.6 & 46.8 & 30.2 & \textbf{54.4} & \textbf{74.5} \\ \midrule
 & \multicolumn{6}{c }{\textbf{top 20}} & \multicolumn{6}{c}{\textbf{top 50}} \\ \midrule
\textbf{BM25} & 77.0 & 49.6 & 60.3 & 37.7 & 28.5 & 69.6 & 46.3 & 66.5 & 54.6 & \textbf{55.1} & 17.6 & 83.2 \\
\textbf{DenseRetr} & \textbf{85.7} & \textbf{56.1} & \textbf{67.8} & 31.4 & 30.4 & 73.5 & 53.8 & 77.0 & 63.4 & 45.7 & 28.0 & 90.8 \\
\textbf{MMR} & 81.4 & 52.7 & 63.9 & 37.1 & 32.2 & 73.2 & 49.8 & 71.4 & 58.7 & 51.3 & 30.3 & 88.7 \\
\textbf{D$k$MIPS} & 74.0 & 49.0 & 59.0 & \textbf{41.8} & 32.1 & 67.7 & 50.3 & 73.7 & 59.8 & 53.7 & 26.5 & 87.7 \\ \midrule
\textbf{NEWSCOPE (GreedySCS)} & 83.4 & 54.3 & 65.8 & 34.9 & 37.8 & 78.4 & 52.2 & 75.0 & 61.6 & 48.7 & \textbf{37.9} & 92.4 \\
\textbf{NEWSCOPE (GreedyPlus)} & 84.1 & 55.8 & 67.1 & 34.9 & \textbf{38.8} & \textbf{84.7} & \textbf{54.1} & \textbf{78.0} & \textbf{63.9} & 47.6 & 32.2 & \textbf{92.7} \\ \bottomrule
\end{tabular}
\caption{Performance on \textbf{LocalNews} benchmark across relevance metrics (P, R, F1) and diversity metrics (D, I, C) at different retrieval depths of top 5, 10, 20, and 50 results.}
\label{tab:localnews}
\end{table*}

\begin{table*}[t]
\centering
\small
\begin{tabular}{@{}c cccccc cccccc@{}}
\toprule
\multirow{3}{*}{\textbf{Model}} 
& \multicolumn{3}{c}{\textbf{Relevancy}} & \multicolumn{3}{c}{\textbf{Diversity}} 
& \multicolumn{3}{c}{\textbf{Relevancy}} & \multicolumn{3}{c}{\textbf{Diversity}} \\
\cmidrule(lr){2-4} \cmidrule(lr){5-7} \cmidrule(lr){8-10} \cmidrule(lr){11-13}
 & \textbf{P} & \textbf{R} & \textbf{F1} & \textbf{D} & \textbf{I} & \textbf{C} & \textbf{P} & \textbf{R} & \textbf{F1} & \textbf{D} & \textbf{I} & \textbf{C} \\ 
 \cmidrule(lr){2-7}\cmidrule(lr){8-13}
 & \multicolumn{6}{c}{\textbf{top 5}} & \multicolumn{6}{c}{\textbf{top 10}} \\ 
 \midrule
\textbf{BM25}           & 92.9 & 10.6 & 19.0 & 25.3 & 45.9 & 25.2 & 91.1 & 20.8 & 33.9 & 29.7 & 36.5 & 40.5 \\
\textbf{DenseRetr}      & \textbf{95.6} & \textbf{10.8} & \textbf{19.4} & 20.7 & 45.3 & 24.6 & \textbf{94.6} & \textbf{21.4} & \textbf{34.9} & 24.6 & 36.2 & 40.0 \\
\textbf{MMR}            & 90.9 & 10.3 & 18.4 & \textbf{35.2} & 52.0 & 28.7 & 90.9 & 20.6 & 33.6 & 32.9 & 40.8 & 45.2 \\
\textbf{D$k$MIPS}       & 90.9 & 10.3 & 18.5 & 31.4 & 49.9 & 24.5 & 90.6 & 20.5 & 33.4 & 32.2 & 38.5 & 39.8 \\ \midrule
\textbf{NEWSCOPE (GreedySCS)}      & 92.2 & 10.4 & 18.7 & 30.8 & 48.2 & 32.3 & 91.8 & 20.8 & 33.9 & 31.5 & 39.1 & 48.1 \\
\textbf{NEWSCOPE (GreedyPlus)} & 91.6 & 10.4 & 18.7 & 33.8 & \textbf{74.4} & \textbf{44.6} & 89.3 & 20.2 & 33.0 & \textbf{34.9} & \textbf{57.4} & \textbf{60.8} \\ \midrule
 & \multicolumn{6}{c}{\textbf{top 20}} & \multicolumn{6}{c}{\textbf{top 50}} \\ \midrule
\textbf{BM25}           & 84.9 & 38.4 & 52.9 & 37.4 & 28.6 & 61.2 & 57.3 & 61.6 & 59.4 & \textbf{55.3} & 18.7 & 82.1 \\
\textbf{DenseRetr}      & \textbf{93.0} & \textbf{42.1} & \textbf{57.9} & 31.3 & 28.8 & 64.6 & \textbf{72.8} & \textbf{77.6} & \textbf{75.2} & 45.7 & 22.8 & 91.8 \\
\textbf{MMR}            & 87.2 & 39.3 & 54.2 & 35.9 & 30.8 & 64.0 & 61.5 & 65.7 & 63.5 & 51.8 & 25.5 & 86.5 \\
\textbf{D$k$MIPS}       & 87.3 & 39.4 & 54.3 & 37.2 & 30.5 & 60.5 & 68.3 & 73.4 & 70.8 & 51.2 & 23.4 & 88.6 \\ \midrule
\textbf{NEWSCOPE (GreedySCS)}      & 88.3 & 39.7 & 54.8 & 35.3 & 31.1 & 67.0 & 62.8 & 66.7 & 64.7 & 49.7 & 24.3 & 90.1 \\
\textbf{NEWSCOPE (GreedyPlus)} & 88.3 & 39.9 & 55.0 & \textbf{35.7} & \textbf{38.3} & \textbf{74.0} & 71.2 & 76.1 & 73.6 & 47.1 & \textbf{26.3} & \textbf{93.1} \\
\bottomrule
\end{tabular}
\caption{Performance on \textbf{DSGlobal} benchmark across relevance metrics (P, R, F1) and diversity metrics (D, I, C) at different retrieval depths of top 5, 10, 20, and 50 results.}
\label{tab:dsglobal}
\end{table*}

\subsection{Global News Benchmark (DSGlobal)}

To assess generalizability beyond local news, we adapt DiverseSumm \citep{DBLP:conf/naacl/HuangLFCJXW24}, a dataset originally designed for news summarization, into a paragraph-level retrieval benchmark. The resulting \textbf{DSGlobal} dataset covers 147 global news events with 7,532 segmented paragraphs. Each paragraph contains 7.5 sentences and 123 words on average. Key statistics comparing LocalNews and DSGlobal are summarized in Table~\ref{tab:stats}.

See Appendix~\ref{sec:appendix_data} for full details on source diversity, data processing, annotation procedures, corresponding prompt templates used and paragraph distribution visualizations.

\section{Evaluation}

\subsection{Experimental Setup}

Given a news headline as the query, the system retrieves a ranked list of the top \( k \) paragraphs from the entire news corpus. Paragraphs unrelated to the event, either from irrelevant reports or different events, are treated as negatives. Stage I retrieves the top 100 relevant paragraphs as candidates for re-ranking. In Stage II, while the cluster coverage threshold offers a flexible stopping condition during greedy selection, we follow standard retrieval evaluation protocols and report fixed top-\(k\) results with \(k \in \{5, 10, 20, 50\}\). Paragraphs are ranked based on their selection order. The threshold remains tunable for customized coverage needs. $\lambda$ is set to be 0.5 for all reported results, as it provides a reasonable trade-off between relevance and diversity.  Results of tuning $\lambda$ across a wider range are reported in Appendix~\ref{app:lambda}.

\subsection{Evaluation Metrics}
We evaluate systems on both standard relevance metrics and novel metrics designed to capture fine-grained semantic diversity.

\paragraph{Precision (P), Recall (R), and F1-score (F1)} Standard metrics for assessing the correctness and completeness of retrieved paragraphs.

\paragraph{Average Pairwise Distance (D)}  
To assess internal diversity, we compute the average pairwise cosine distance among retrieved paragraph embeddings:
\begin{align*}
D = \frac{2}{k(k-1)} \sum_{i=1}^{k} \sum_{j=i+1}^{k} \left( 1 - \cos(\mathbf{p}_i, \mathbf{p}_j) \right)
\end{align*}
Higher values indicate greater semantic variation.

\paragraph{Positive Cluster Coverage (C)}  
We define diversity in terms of sentence-level semantics. Each sentence is assigned to a semantic cluster (Section 4.2.1), and each cluster represents a unique perspective in news reporting. \(C\) measures the proportion of these clusters covered by the top-\(k\) retrieved paragraphs:
\begin{align*}
C = \frac{\text{\# Covered Clusters}}{\text{\# Total Clusters in Relevant Paragraphs}}
\end{align*}
This metric captures diversity through fine-grained semantic recall and avoids rewarding redundancy.

\paragraph{Information Density Ratio (I)}  
To evaluate informativeness, we compute the average number of unique clusters represented in the retrieved paragraphs:
\begin{align*}
I = \frac{\text{\# Covered Clusters}}{\text{\# Total Sentences in Retrieved Paragraphs}}
\end{align*}
This penalizes repetition and favors concise, information-rich paragraphs.

Together, \(C\), \(D\), and \(I\) provide a multi-dimensional assessment of semantic coverage, viewpoint diversity, and content efficiency.

\subsection{Baselines}

We compare our framework against strong retrieval methods with different strategies for balancing relevance and diversity.

\noindent\textbf{BM25} A sparse retrieval model based on term frequency and inverse document frequency \cite{robertson1995okapi}, highly effective for relevance matching but lacking diversity modeling.

\noindent\textbf{DenseRetr (Dense Retrieval)}  A multilingual dense retriever \texttt{bilingual-embedding-large} \cite{thakur2020augmented}, fine-tuned on STS and NLI tasks. It is used in Stage I of our framework, strong on relevance but diversity-agnostic.

\noindent\textbf{MMR (Maximal Marginal Relevance)}  
A classical method that promotes diversity by selecting documents that maximize relevance to the query while minimizing redundancy with already selected items \cite{carbonell1998use}. 

\noindent\textbf{D\(k\)MIPS (Diversity-aware \(k\)-Maximum Inner Product Search)}  
A method that jointly optimizes for relevance and diversity by maximizing similarity to the query and minimizing pairwise similarity among retrieved items \cite{DBLP:journals/corr/abs-2402-13858}. 

All methods use the same input query (event headline) and top-\(k\) output size. The formal scoring functions for baselines BM25, MMR, and D\(k\)MIPS are provided in Appendix~\ref{sec:appendix_baseline_equations}.

\section{Results \& Analysis}
\subsection{Main Results}

\begin{figure}[t]
    \centering
    \begin{subfigure}[b]{\linewidth}
        \centering
        \includegraphics[width=\linewidth]{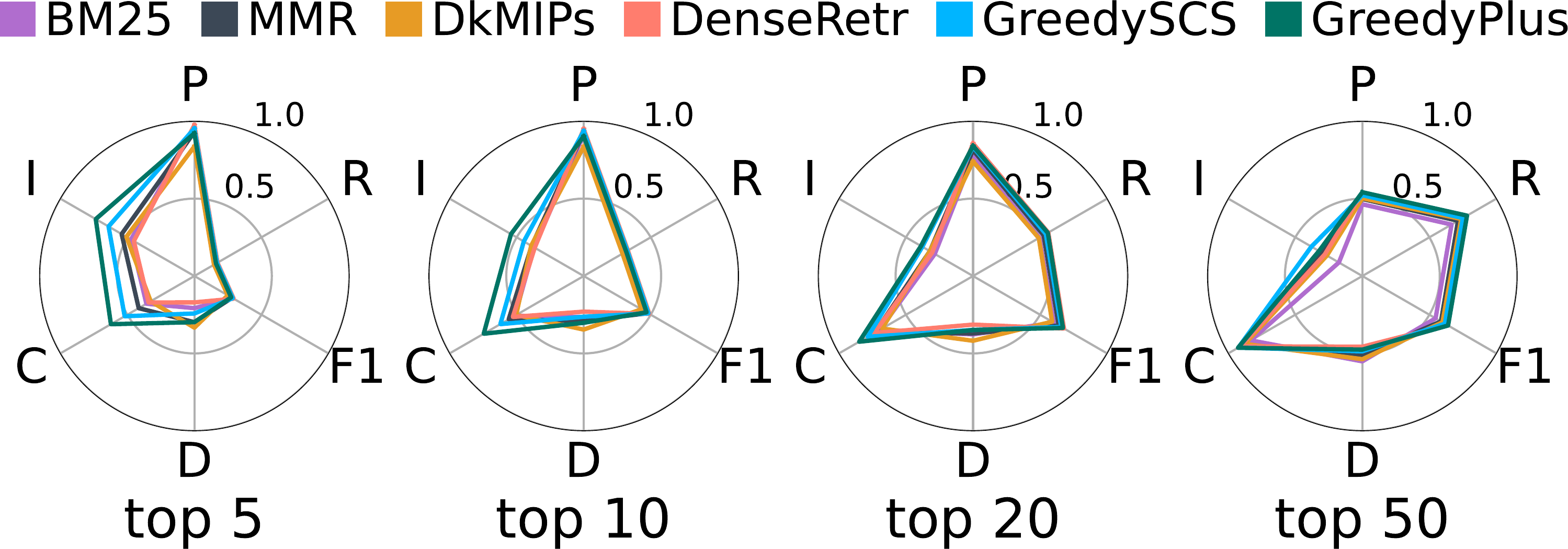}
        \caption{LocalNews}
    \end{subfigure}
    \hfill
    \begin{subfigure}[b]{\linewidth}
        \centering
        \includegraphics[width=\linewidth]{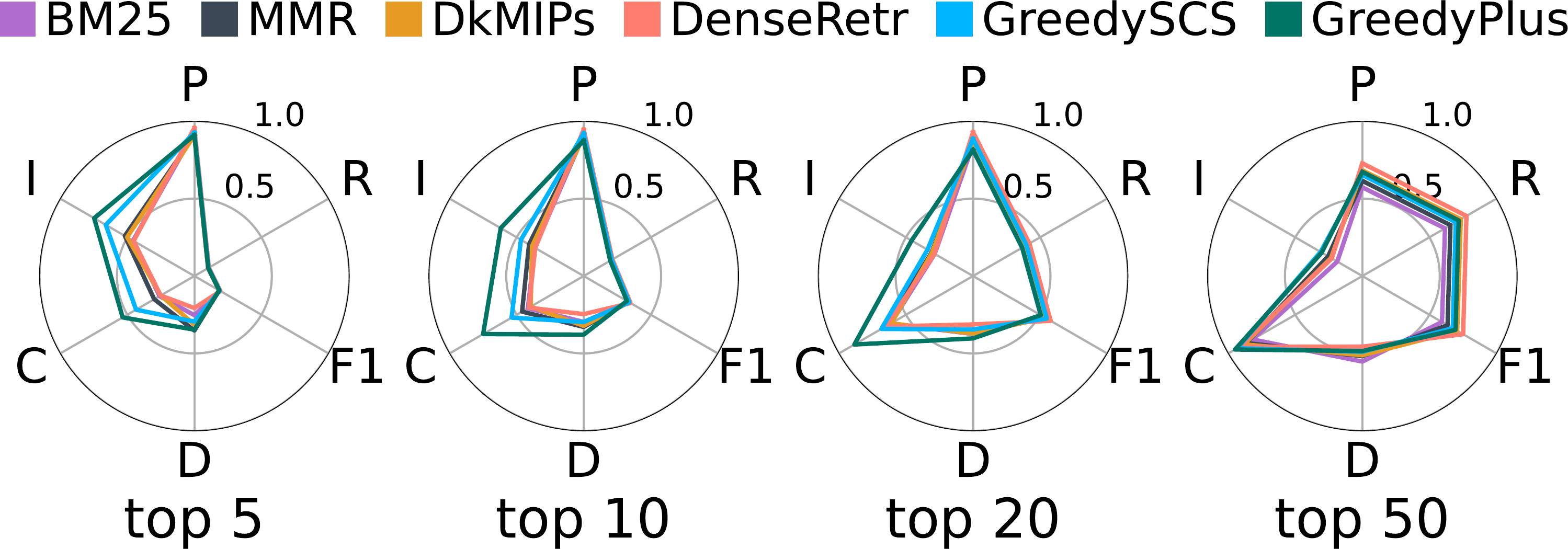}
        \caption{DSGlobal}
    \end{subfigure}
            \caption{Visualization of model performance across different retrieval depth levels.}
    \label{fig:performance}
\end{figure}

We report results on the LocalNews and DSGlobal benchmarks in Table~\ref{tab:localnews} and Table~\ref{tab:dsglobal}, evaluated across four retrieval depths (top 5, 10, 20, and 50) using standard relevance metrics, Precision (P), Recall (R), and F1, and diversity-oriented metrics, Average Pairwise Distance (D), Information Density Ratio (I), and Positive Cluster Coverage (C).

\begin{table}[t]
\centering
\small
\begin{tabular}{@{}c p{3.1cm} p{3.1cm}@{}}
\toprule
\textbf{Rank} & \textbf{DenseRetr} & \textbf{NEWSCOPE} \\ \midrule
\textbf{1} & Altman expresses excitement and concern. & Altman: benefits outweigh risks. \\
\textbf{2} & Altman: benefits outweigh risks. & Russia’s AI interest; hallucinations. \\
\textbf{3} & Altman expresses excitement and concern. & ChatGPT adoption and benchmarks. \\
\textbf{4} & Altman: benefits outweigh risks. & Altman vs. Virginia Governor on AI in education. \\
\textbf{5} & Altman expresses excitement and concern. & Criticism: ChatGPT unreliable and uncreative. \\
\bottomrule
\end{tabular}
\caption{Top-5 retrieved paragraphs (summarized) under dense retrieval and proposed diverse retrieval.}
\label{tab:qualitative_summary}
\end{table}

\begin{table*}[t]
\centering
\small
\setlength{\tabcolsep}{4.5pt} 
\begin{tabular}{@{}c@{\hskip 6pt}cccc@{\hskip 4pt}cccc@{\hskip 4pt}cccc@{\hskip 4pt}cccc@{}}
\toprule
\multirow{3}{*}{\textbf{Model}} & \multicolumn{16}{c}{\textbf{LocalNews}} \\ \cmidrule{2-17}
& \multicolumn{4}{c}{\textbf{Top 5}} & \multicolumn{4}{c}{\textbf{Top 10}} & \multicolumn{4}{c}{\textbf{Top 20}} & \multicolumn{4}{c}{\textbf{Top 50}} \\
\cmidrule(r){2-5} \cmidrule(lr){6-9} \cmidrule(lr){10-13} \cmidrule(l){14-17}
& \textbf{F1} & \textbf{D} & \textbf{I} & \textbf{C} & \textbf{F1} & \textbf{D} & \textbf{I} & \textbf{C} & \textbf{F1} & \textbf{D} & \textbf{I} & \textbf{C} & \textbf{F1} & \textbf{D} & \textbf{I} & \textbf{C} \\
\midrule
\textbf{GreedyPlus} 
& 27.5 & 29.8 & 73.6 & 62.2 
& 46.8 & 30.2 & 54.4 & 74.5 
& 67.1 & 34.9 & 38.8 & 84.7 
& 63.9 & 47.6 & 32.2 & 92.7 \\

\textbf{w/o diversity term} 
& \textbf{29.3} & 17.7 & 45.1 & 35.8 
& \textbf{48.8} & 24.1 & 36.1 & 52.8 
& \textbf{68.3} & 33.3 & 30.1 & 75.0 
& \textbf{64.4} & 47.2 & 26.7 & 90.9 \\

\textbf{w/o relevance term} 
& 25.2 & \textbf{34.4} & \textbf{82.3} & \textbf{64.9} 
& 32.0 & \textbf{42.9} & \textbf{76.0} & \textbf{80.5} 
& 30.7 & \textbf{51.0} & \textbf{72.2} & \textbf{87.4} 
& 57.4 & \textbf{49.2} & \textbf{42.7} & \textbf{93.1} \\ \midrule

& \multicolumn{16}{c}{\textbf{DSGlobal}} \\ \midrule

\textbf{GreedyPlus} 
& 18.7 & 33.8 & 74.4 & 44.6 
& 33.0 & 34.9 & 57.4 & 60.8 
& 55.0 & 35.7 & 38.3 & 74.0 
& 73.6 & 47.1 & 26.3 & \textbf{93.1} \\

\textbf{w/o diversity term} 
& \textbf{19.4} & 22.2 & 46.6 & 24.3 
& \textbf{34.6} & 26.9 & 37.2 & 39.9 
& \textbf{56.3} & 33.5 & 29.8 & 64.1 
& \textbf{74.5} & 46.7 & 22.8 & 91.7 \\

\textbf{w/o relevance term} 
& 18.5 & \textbf{35.1} & \textbf{76.9} & \textbf{45.1} 
& 30.4 & \textbf{41.5} & \textbf{66.5} & \textbf{63.4} 
& 35.5 & \textbf{51.3} & \textbf{58.4} & \textbf{75.1} 
& 59.1 & \textbf{52.3} & \textbf{36.0} & 90.3 \\ 
\bottomrule
\end{tabular}
\caption{Ablation study of \textbf{NEWSCOPE (GreedyPlus)} on \textbf{LocalNews} and \textbf{DSGlobal.}}
\label{tab:ablation_sum}
\end{table*}

\vspace{0.3em}
\noindent\textbf{Relevance-focused baselines.} Among retrieval methods, BM25 and DenseRetr achieve strong relevance scores. DenseRetr yields the highest precision, recall, and F1, demonstrating effective semantic matching. However, it consistently underperforms on diversity metrics (D, I, and C), reflecting its tendency to retrieve redundant content with limited perspective variation.

\vspace{0.3em}
\noindent\textbf{Diversity-aware baselines.} MMR and D\(k\)MIPS improve diversity through novelty-aware selection. MMR yields modest gains in diversity, especially at shallow depths. D\(k\)MIPS achieves stronger improvements in diversity, but sacrifices relevance performance, leading to lower overall F1.

\vspace{0.3em}
\noindent\textbf{Effectiveness of \textbf{NEWSCOPE}.} As shown in Figure~\ref{fig:performance}, both variants of \textbf{NEWSCOPE} substantially outperform baselines on diversity metrics. GreedySCS achieves a strong balance, enhancing Positive Cluster Coverage (C) with minimal precision loss. GreedyPlus further boosts diversity, attaining the highest scores on Information Density Ratio (I) and C across nearly all settings. Performance trends are consistent across LocalNews and DSGlobal, confirming the robustness of our framework.

\vspace{0.3em}
\noindent\textbf{Depth-wise trade-offs.} Figure~\ref{fig:performance} shows that at shallow depths (top 5 and 10), GreedySCS and GreedyPlus achieve substantially higher diversity scores (D, I, C) while maintaining strong relevance, highlighting their ability to uncover diverse perspectives early. At deeper depths, all methods gain higher coverage as more paragraphs are retrieved, while \textbf{NEWSCOPE} achieves the best diversity.

Overall, the results demonstrate that \textbf{NEWSCOPE} achieves the best balance of relevance and diversity, enabling more comprehensive event understanding.

\subsection{Qualitative Analysis}

We compare relevance-based retrieval (DenseRetr) with our diversity-aware re-ranking (GreedyPlus) to illustrate improvements in factual coverage and reduction in redundancy.

As shown in Table~\ref{tab:qualitative_summary}, Given the query ``\textit{OpenAI CEO Sam Altman warns of AI risks, calling for careful regulation amid global competition}'', DenseRetr returns near-duplicate content focused on Altman’s interview, while GreedyPlus retrieves a wider range of perspectives, including global competition, societal risks, policy debates, and criticisms of generative AI. This illustrates how sentence-level modeling and cluster-based re-ranking enable broader yet relevant coverage. Full examples appear in Appendix~\ref{sec:appendix_qualitative}.

\subsection{Ablation Study}

We conduct an ablation study to assess the contributions of the relevance and diversity components in GreedyPlus. Table~\ref{tab:ablation_sum} reports summarized results on both datasets; full tables are provided in Appendix~\ref{sec:appendix_ablation_result}.

Removing the diversity term reduces the method to a relevance-based reranking. While slightly improveing F1, it significantly reduces all diversity metrics, leading to redundant results. In contrast, removing the relevance term maximizes diversity but harms F1, selecting loosely relevant or off-topic content. GreedyPlus consistently achieves the best balance across retrieval depths, demonstrating the importance of jointly modeling relevance and diversity. Qualitative examples for the ablation study are provided in Appendix~\ref{sec:appendix_ablation}.

\section{Conclusion}

In this work, we propose \textbf{NEWSCOPE}, a two-stage framework for diverse news retrieval that enhances event coverage by modeling semantic variation with fine-grained granularity. By integrating sentence-level clustering with interpretable greedy re-ranking, our method surfaces complementary perspectives without sacrificing relevance. We propose three novel diversity metrics, construct two benchmarks spanning local and global contexts, and show consistent gains over strong baselines. By revealing underrepresented viewpoints, this work highlights the importance of diversity-aware retrieval in mitigating redundancy and promoting unbiased access to information. Future directions include domain generalization and integration with fact verification for more trustworthy information retrieval.

\newpage
\section*{Limitations}

While our approach improves fine-grained interpretable diversity in news retrieval, several limitations remain. First, our selection mechanism optimizes for semantic variation at the sentence level but does not explicitly control for factual consistency across retrieved paragraphs. As a result, the retrieved set may contain conflicting narratives. Future extensions could incorporate source credibility to better balance diversity, accuracy, and contextual coherence. Second, current scoring treats clusters independently. Though OPTICS helps separate dense regions, explicitly modeling inter-cluster similarity is a promising direction.

Finally, although we focus on the news domain where divergent perspectives naturally occur, our framework is domain-agnostic. Future work could explore its adaptability to domains such as finance, science, or law, where diversity takes different forms and factual consistency is often prioritized over viewpoint variation.

\section*{Acknowledgments}
This research is supported by the Ministry of Education, Singapore, under its MOE AcRF TIER 3 Grant (MOE-MOET32022-0001).

\bibliography{custom}

\appendix
\section{Data Construction Details: Source Diversity and Annotation Process}
\label{sec:appendix_data}

\subsection{Data Sources}
\textbf{LocalNews} is curated to support sentence-level diversity modeling in event-centric retrieval. It focuses on localized events and is constructed using Google News’ ``Full Coverage'' feature, which aggregates articles from various sources reporting on the same event. This method helps mitigate source bias by incorporating a broad spectrum of viewpoints.

We analyzed the collected sources and found a mix of:
\begin{itemize}
    \item \textbf{Mainstream outlets} (e.g., major national newspapers and broadcasters),
    \item \textbf{Independent media platforms},
    \item \textbf{Domain-specific publishers} (e.g., business- or health-focused),
    \item \textbf{Regional and international media}, including nearby countries and global organizations.
\end{itemize}

To ensure consistency and avoid personalization bias, we accessed ``Full Coverage'' pages from varying IP addresses (via VPN), devices, login states (logged-in vs. incognito), and locations. The resulting content remained stable across conditions.

\subsection{Data Processing}
We applied the following preprocessing steps to construct the benchmark:
\begin{itemize}
    \item \textbf{Event De-duplication}: Articles reporting on the same incident were grouped and deduplicated to ensure a clean set of distinct events.
    \item \textbf{Paragraph Segmentation}: Articles were segmented into coherent paragraphs, capped at 512 tokens, which aligns with the input limits of modern encoding models. Paragraph boundaries are preserved wherever possible; overly long paragraphs are truncated at sentence boundaries. The final dataset contains 5,296 paragraphs, averaging 7.1 sentences and 124.9 words each.
\end{itemize}

\subsection{Annotation Procedure}
We adopt an event-centric labeling strategy:
\begin{itemize}
    \item \textbf{Query Generation}: One-sentence abstractive summaries were generated using GPT-4o-mini to to represent each event. These summaries served as queries for retrieval, simulating user searches for event-related information. To ensure high-quality queries, we iteratively refined summaries where the positive rate of relevance labels did not meet a predefined threshold with up to five rounds.
    \item \textbf{Relevance Labeling}: Each paragraph was annotated for relevance to the event summary. We used majority voting across three rounds of annotation, achieving an inter-annotation agreement rate of 98.7\%. Paragraphs from articles outside the event’s full coverage were auto-labeled as irrelevant.
\end{itemize}

\subsection{Prompt Templates}

We employ LLM prompting to assist in dataset labelling for (1) event headline generation and (2) paragraph relevance annotation. Below are the two prompt templates:

\begin{tcolorbox}[colback=gray!5, colframe=gray!50!black, title=Prompt: Generate Summarized Headline]

Read the following news articles discussing the same event. Produce a very brief headline that summarizes the general aspects of the event \textbf{in one sentence}. Include names of key entities (e.g., people, countries, companies, or products) when relevant to help refer to the event. Only output the headline, without any explanation.

\texttt{==========}

Article 0: \{content of Article 0\}

Article 1: \{content of Article 1\}

Article 2: \{content of Article 2\}

... (repeat until max\_token is reached)

\texttt{==========}

\textbf{Reminder:} Your answer must be a one-sentence headline focusing on the main event and important entities. Do not include additional explanation.
\end{tcolorbox}

\newpage

\begin{tcolorbox}[colback=gray!5, colframe=gray!50!black, title=Prompt: Relevance Classification]
Please assess the relevance between the following news headline and paragraph. \textbf{If the paragraph discusses the event, provides background, or describes consequences}, output \texttt{1}. If it is not relevant, output \texttt{0}. Output only \texttt{1} or \texttt{0}, without any additional text.

[Headline begins] "\{headline\}" [Headline ends]

[Paragraph begins] "\{paragraph\}" [Paragraph ends]

Your output:

\end{tcolorbox}

\subsection{Paragraph Distribution}

\begin{figure}[h]
    \centering
    \begin{subfigure}[b]{0.49\linewidth}
        \centering
        \includegraphics[width=\linewidth]{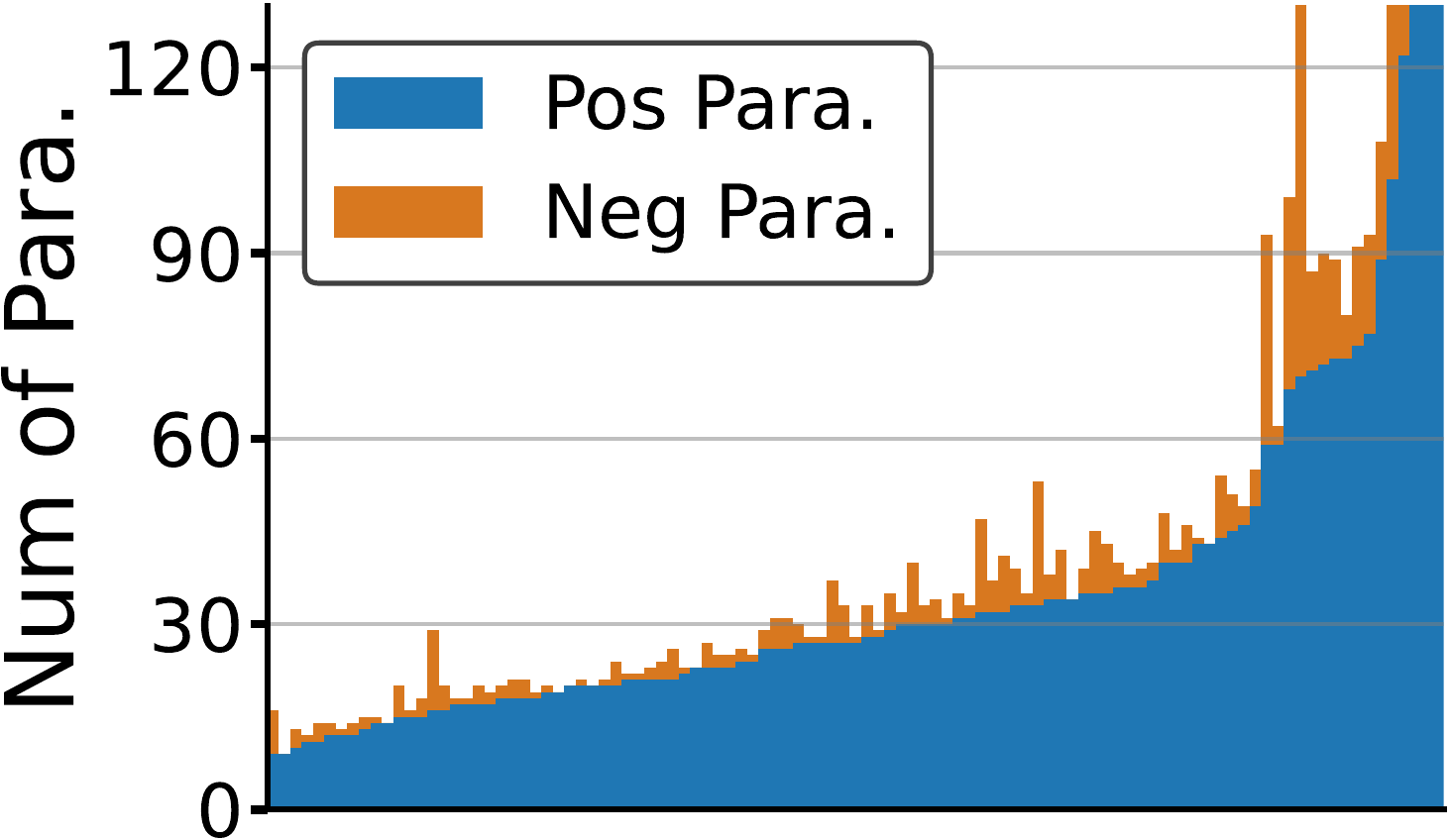}
        \caption{LocalNews}
    \end{subfigure}
    \hfill
    \begin{subfigure}[b]{0.49\linewidth}
        \centering
        \includegraphics[width=\linewidth]{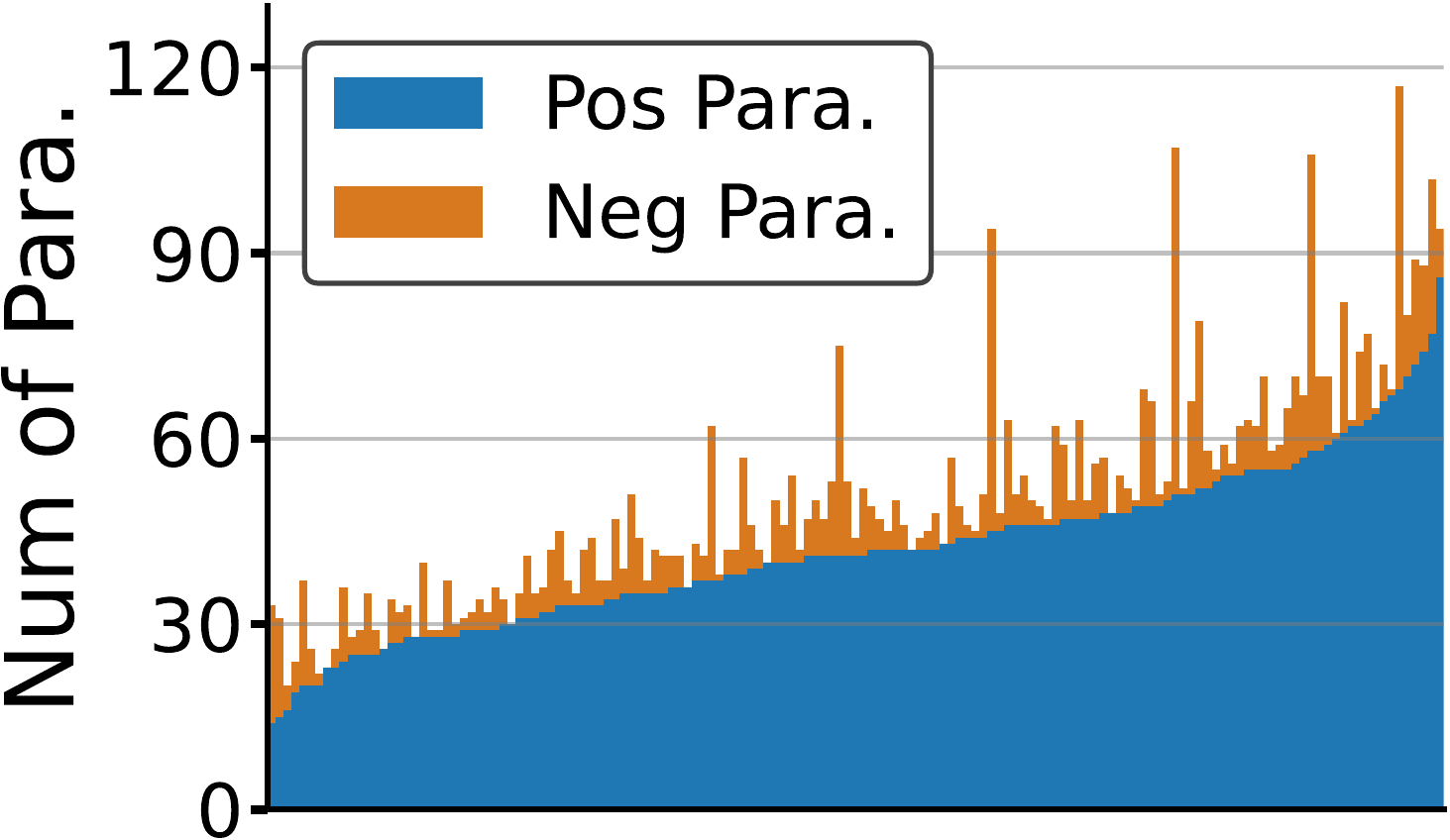}
        \caption{DSGlobal}
    \end{subfigure}
    \caption{Positive vs. negative paragraph counts per event.}
    \label{fig:pos_neg_dist}
\end{figure}

\begin{figure}[h]
    \centering
    \begin{subfigure}[b]{0.49\linewidth}
        \centering
        \includegraphics[width=\linewidth]{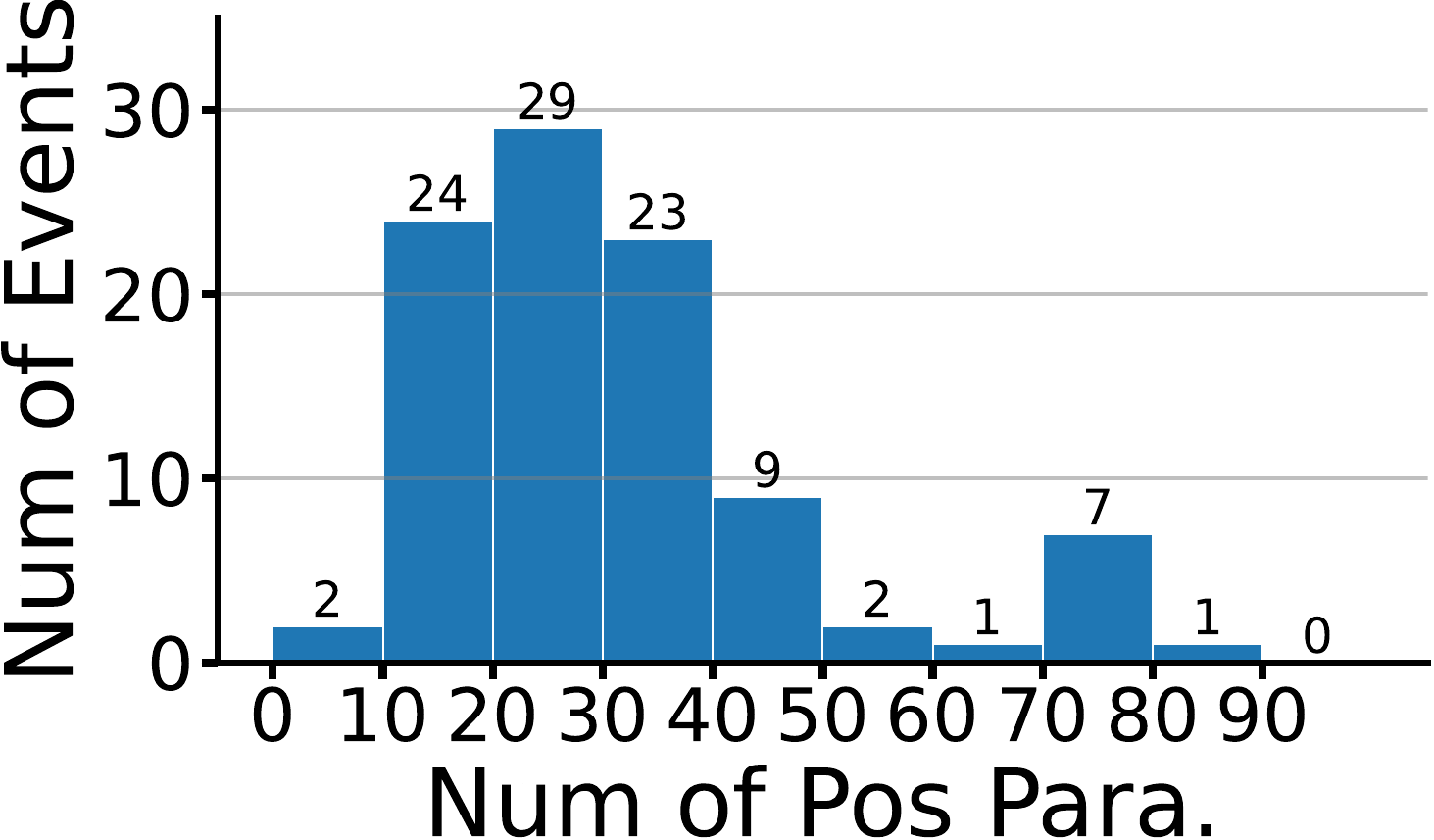}
        \caption{LocalNews}
    \end{subfigure}
    \hfill
    \begin{subfigure}[b]{0.49\linewidth}
        \centering
        \includegraphics[width=\linewidth]{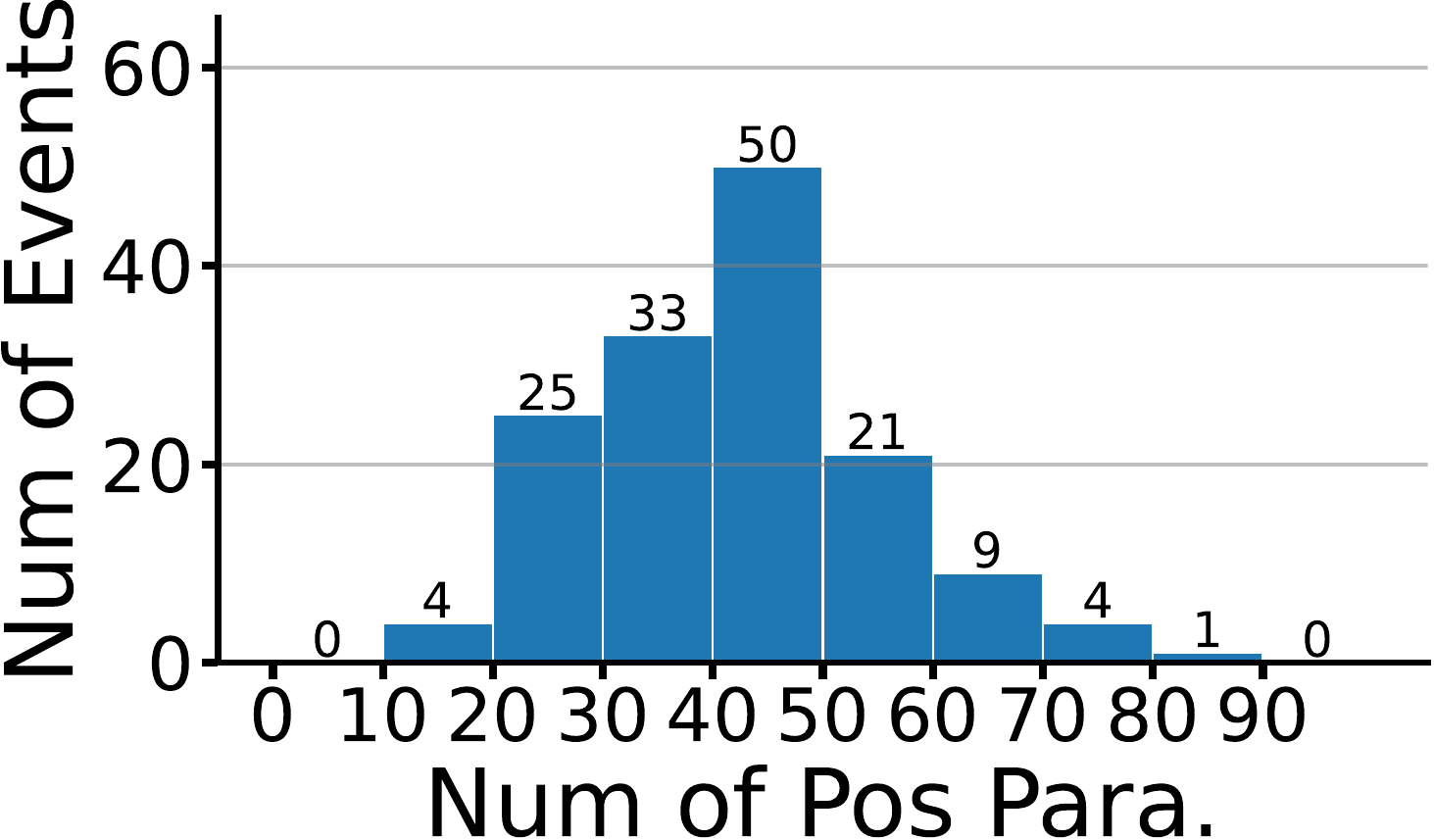}
        \caption{DSGlobal}
    \end{subfigure}
    \caption{Event distribution by number of positive paragraphs.}
    \label{fig:pos_range}
\end{figure}

Figure~\ref{fig:pos_neg_dist} compares the distribution of positive and negative paragraphs per event, while Figure~\ref{fig:pos_range} shows the variation in the number of positive paragraphs across events in both datasets.

\section{Example Clusters}
\label{app:5cluster}

To illustrate that sentence clusters serve as effective proxies for event aspects, we present the first five clusters derived from an example.

\paragraph{Cluster 1}
\begin{itemize}
    \item At about 1am, Scoot staff started distributing mineral water and bread to passengers, but still did not allow them to board the plane.
    \item Scoot staff apparently distributed mineral water and bread to passengers but they were still unable to board the plane.
    \item While the Scoot staff handed out bottled water and bread, they still didn’t allow us to board.
\end{itemize}

\paragraph{Cluster 2}
\begin{itemize}
    \item Many passengers kept asking them what the problem was but the staff kept saying it was due to the weather.
    \item Many passengers kept asking them what the problem was and whether there was something wrong with the plane, but they kept saying it was due to the weather.
    \item Many passengers kept asking what the issue was, but the staff insisted it was due to the weather, he said.
\end{itemize}

\paragraph{Cluster 3}
\begin{itemize}
    \item It was only at 2.30am when the staff finally allowed passengers to board the plane, and Mr Lin fell asleep immediately when he got in his seat.
    \item Only at 2:30am did the staff allow passengers to board the plane.
\end{itemize}

\paragraph{Cluster 4}
\begin{itemize}
    \item However when he woke up half an hour later, the plane had not taken off yet.
    \item Lin shared that he fell asleep immediately after getting onto the plane but when he woke up half an hour later, they had not taken off yet.
    \item At 2.30am, passengers were finally given the green light to board and Lin said he fell asleep shortly after boarding but woke up 30 minutes later to find the plane still at the runway.
\end{itemize}

\paragraph{Cluster 5}
\begin{itemize}
    \item The captain made announced that he could not start the engine and the plane had to return to the hangar, so everyone had to disembark.
    \item The plane’s captain then apparently announced that the plane had to return to the hangar due to the engine not starting, and asked the passengers to disembark.
    \item The captain then informed us via the PA system that the plane’s engine couldn’t start and had to be towed back to the hangar and told us to disembark from the plane.
\end{itemize}

\section{Efficiency Analysis}
\label{sec:appendix_efficiency}

While our framework introduces a second-stage re-ranking module, it is designed with practical efficiency in mind. In Stage I, we retrieve the top 100 paragraphs using a dense retriever, which acts as a lightweight filter. This greatly reduces the candidate pool for Stage II, where sentence-level clustering and re-ranking are applied.

Retrieving 50 final items from this pool already covers over 80\% of relevant clusters on average (as shown in main results), confirming that 100 candidates provide sufficient scope for effective and diverse re-ranking.

We benchmarked the average runtime across 100 events to assess the end-to-end efficiency of the \textbf{NEWSCOPE} pipeline:

\begin{table}[h]
\centering
\begin{tabular}{@{}l|c@{}}
\toprule
\textbf{Step} & \textbf{Time (s)} \\
\midrule
Dense retrieval & 0.025 \\
Sentence encoding \& clustering & 0.84 \\
Re-ranking (Greedy Selection) & 0.51 \\
\midrule
\textbf{Total} & \textbf{1.175} \\
\bottomrule
\end{tabular}
\caption{Average runtime per event across 100 examples.}
\end{table}

This breakdown demonstrates that our full pipeline operates within ~1.2 seconds per event, confirming its scalability for real-world diverse news retrieval applications.

\section{Baseline Scoring Functions}
\label{sec:appendix_baseline_equations}

\paragraph{BM25}  
BM25 \cite{robertson1995okapi} ranks documents based on term frequency and inverse document frequency, balancing relevance with length normalization. The scoring function is:
\begin{align*}
\text{BM25}(Q, D) &= \sum_{t \in Q} \log\frac{N - n_t + 0.5}{n_t + 0.5} \\
& \cdot \frac{(k_1 + 1)f_{t,D}}{k_1(1 - b + b \cdot \frac{|D|}{\text{avgdl}}) + f_{t,D}}
\end{align*}

\paragraph{MMR (Maximal Marginal Relevance)}  
MMR \cite{carbonell1998use} balances relevance and diversity by selecting the document \( D_i \) that maximizes:
\begin{align*}
\text{MMR}(D_i, Q) &= \arg\max_{D_i \in \mathcal{D}\setminus S} \Big( \lambda \cdot \text{Sim}(D_i, Q) \\
& - (1 - \lambda) \cdot \max_{D_j \in S} \text{Sim}(D_i, D_j) \Big)
\end{align*}
where \( \lambda \in [0,1] \) controls the trade-off.

\paragraph{D\(k\)MIPS}  
D\(k\)MIPS \cite{DBLP:journals/corr/abs-2402-13858} formulates diversity-aware retrieval as a penalized inner product optimization:
\begin{align*}
f(S) := \lambda \sum_{\mathbf{p} \in S} \langle \mathbf{p}, \mathbf{q} \rangle - \frac{2\mu (1 - \lambda)}{k(k-1)} \sum_{\mathbf{p}, \mathbf{p'} \in S} \langle \mathbf{p}, \mathbf{p'} \rangle
\end{align*}
This encourages both high relevance to the query \( \mathbf{q} \) and low redundancy among retrieved paragraphs \( \mathbf{p} \in S \).

\section{Qualitative Analysis Example: Full Text}
\label{sec:appendix_qualitative}

The following table contains the retrieval results of news headline: ``\textit{OpenAI CEO Sam Altman expresses concerns about AI's risks while highlighting ChatGPT's transformative potential and the need for careful regulation amidst competition from global players like China and Russia}.''

These results illustrate how our method surfaces complementary angles, including technological, political, educational, and critical, enhancing users’ understanding of complex events beyond repetitive core quotes.

\begin{table*}[htbp]
\centering 
\caption{Comparison of Top-5 Results from Relevance Retrieval and Diverse Retrieval}
\label{tab:my_appendix_table_booktabs_final}
\small
\newcolumntype{N}{>{\centering\arraybackslash}p{0.4cm}} 
\begin{tabularx}{\linewidth}{@{} N >{\RaggedRight\arraybackslash}X >{\RaggedRight\arraybackslash}X @{}}
\toprule 
\addlinespace[2pt]
& \multicolumn{1}{c}{\textbf{Relevance Retrieval: DenseRetr}} & \multicolumn{1}{c}{\textbf{Diverse Retrieval: NEWSCOPE (GreedyPlus)}} \\
\addlinespace[2pt]

\addlinespace[2pt]
\multicolumn{1}{c}{\textbf{1}}  & Sam Altman, co-founder and chief executive officer of OpenAI Inc., speaks during TechCrunch Disrupt 2019 in San Francisco, California, on Thursday, Oct. 3, 2019.OpenAI CEO Sam Altman said in a recent interview with ABC News that he's a "little bit scared" of artificial intelligence technology and how it could affect the workforce, elections and the spread of disinformation.OpenAI developed the ChatGPT bot, which creates human-like answers to questions and ignited a new AI craze."I think people really have fun with [ChatGPT]," Altman said in the interview.But his excitement over the transformative potential of AI technology, which Altman said will eventually reflect "the collective power, and creativity, and will of humanity," was balanced by his concerns about "authoritarian regimes" developing competing AI technology."We do worry a lot about authoritarian governments developing this," Altman said. Overseas governments have already begun to bring competing AI technology to market. & Among the concerns of the destructive capabilities of this technology is the replacement of jobs. Altman says this will likely replace some jobs in the near future, and worries how quickly that could happen."I think over a couple of generations, humanity has proven that it can adapt wonderfully to major technological shifts," Altman said. "But if this happens in a single-digit number of years, some of these shifts ... That is the part I worry about the most."But he encourages people to look at ChatGPT as more of a tool, not as a replacement. He added that "human creativity is limitless, and we find new jobs. We find new things to do."ABC News OpenAI CEO Sam Altman speaks with ABC News, Mar. 15, 2023.The ways ChatGPT can be used as tools for humanity outweigh the risks, according to Altman."We can all have an incredible educator in our pocket that's customized for us, that helps us learn," Altman said. "We can have medical advice for everybody that is beyond what we can get today.” \\
\addlinespace 
\multicolumn{1}{c}{\textbf{2}}  & Among the concerns of the destructive capabilities of this technology is the replacement of jobs. Altman says this will likely replace some jobs in the near future, and worries how quickly that could happen."I think over a couple of generations, humanity has proven that it can adapt wonderfully to major technological shifts," Altman said. "But if this happens in a single-digit number of years, some of these shifts ... That is the part I worry about the most."But he encourages people to look at ChatGPT as more of a tool, not as a replacement. He added that "human creativity is limitless, and we find new jobs. We find new things to do."ABC News OpenAI CEO Sam Altman speaks with ABC News, Mar. 15, 2023.The ways ChatGPT can be used as tools for humanity outweigh the risks, according to Altman."We can all have an incredible educator in our pocket that's customized for us, that helps us learn," Altman said. "We can have medical advice for everybody that is beyond what we can get today.” &  Countries such as Russia have shown an interest in pursuing AI. Russian President Vladimir Putin told students in 2017 that whoever led the AI race would most likely "rule the world." GPT-4 has safeguards to protect users from engaging in illicit conduct, such as seeking information about how to construct bombs.Altman also noted the software's inability to fact-check. "The thing that I try to caution people the most is what we call the 'hallucinations problem,'" he said. "The model will confidently state things as if they were facts that are entirely made up." OpenAI is attempting to counter this problem by having the bot use deductive reasoning rather than memorization, allowing it to process statements in real time.OpenAI launched the latest version of its software, GPT-4, on Tuesday. The bot has a faster response rate and can process image prompts.CLICK HERE TO READ MORE FROM THE WASHINGTON EXAMINER \\
\addlinespace
\multicolumn{1}{c}{\textbf{3}}  & ABC News(NEW YORK) -- The CEO behind the company that created ChatGPT believes artificial intelligence technology will reshape society as we know it. He believes it comes with real dangers, but can also be "the greatest technology humanity has yet developed" to drastically improve our lives."We've got to be careful here," said Sam Altman, CEO of OpenAI. "I think people should be happy that we are a little bit scared of this."Altman sat down for an exclusive interview with ABC News' chief business, technology and economics correspondent Rebecca Jarvis to talk about the rollout of GPT-4 -- the latest iteration of the AI language model.In his interview, Altman was emphatic that OpenAI needs both regulators and society to be as involved as possible with the rollout of ChatGPT -- insisting that feedback will help deter the potential negative consequences the technology could have on humanity. He added that he is in "regular contact" with government officials. & ChatGPT is an AI language model, the GPT stands for Generative Pre-trained Transformer.Released only a few months ago, it is already considered the fastest-growing consumer application in history. The app hit 100 million monthly active users in just a few months. In comparison, TikTok took nine months to reach that many users and Instagram took nearly three years, according to a UBS study.Watch the exclusive interview with Sam Altman on "World News Tonight with David Muir" at 6:30 p.m. ET on ABC.Though "not perfect," per Altman, GPT-4 scored in the 90th percentile on the Uniform Bar Exam. It also scored a near-perfect score on the SAT Math test, and it can now proficiently write computer code in most programming languages.GPT-4 is just one step toward OpenAI's goal to eventually build Artificial General Intelligence, which is when AI crosses a powerful threshold which could be described as AI systems that are generally smarter than humans. \\
\addlinespace
\bottomrule
\addlinespace
\multicolumn{3}{r}{\textit{Table continued in next page.}} \\*

\end{tabularx}
\end{table*}

\newpage
\begin{table*}[htbp]
\centering

\small
\newcolumntype{N}{>{\centering\arraybackslash}p{0.4cm}}

\begin{tabularx}{\linewidth}{@{} N >{\RaggedRight\arraybackslash}X >{\RaggedRight\arraybackslash}X @{}}
\toprule 

\addlinespace[2pt]
& \multicolumn{1}{c}{\textbf{Relevance Retrieval: DenseRetr}} & \multicolumn{1}{c}{\textbf{Diverse Retrieval: NEWSCOPE (GreedyPlus)}} \\
\addlinespace[2pt]

\addlinespace[2pt]
\multicolumn{1}{c}{\textbf{4}}  & Among the concerns of the destructive capabilities of this technology is the replacement of jobs. Altman says this will likely replace some jobs in the near future, and worries how quickly that could happen."I think over a couple of generations, humanity has proven that it can adapt wonderfully to major technological shifts," Altman said. "But if this happens in a single-digit number of years, some of these shifts ... That is the part I worry about the most."But he encourages people to look at ChatGPT as more of a tool, not as a replacement. He added that "human creativity is limitless, and we find new jobs. We find new things to do."OpenAI CEO Sam Altman speaks with ABC News, Mar. 15, 2023. ABC NewsThe ways ChatGPT can be used as tools for humanity outweigh the risks, according to Altman."We can all have an incredible educator in our pocket that's customized for us, that helps us learn," Altman said. "We can have medical advice for everybody that is beyond what we can get today.”   & He continued: "It is going to eliminate a lot of current jobs, that's true. We can make much better ones. The reason to develop AI at all, in terms of impact on our lives and improving our lives and upside, this will be the greatest technology humanity has yet developed."Altman also spoke on the impacts that AI-powered chatbots would have on education and whether it would "increase laziness among students.""Education is going to have to change," the OpenAI CEO said. "But it's happened many other times with technology. When we got the calculator, the way we taught math and what we tested students on totally changed."VIRGINIA GOV. YOUNGKIN SAYS MORE SCHOOLS SHOULD BAN CHATGPTVirginia Gov. Glenn Youngkin announced in March that more school districts should ban AI technologies like ChatGPT.Youngkin said that the goal of education was "to make sure that our kids can think and, therefore, if a machine is thinking for them, then we're not accomplishing our goal.” \\
\addlinespace
\multicolumn{1}{c}{\textbf{5}}  & OpenAI CEO Sam Altman believes artificial intelligence has incredible upside for society, but he also worries about how bad actors will use the technology.In an ABC News interview this week, he warned "there will be other people who don't put some of the safety limits that we put on."OpenAI released its A.I. chatbot ChatGPT to the public in late November, and this week it unveiled a more capable successor called GPT-4.Other companies are racing to offer ChatGPT-like tools, giving OpenAI plenty of competition to worry about, despite the advantage of having Microsoft as a big investor."It's competitive out there," OpenAI cofounder and chief scientist Ilya Sutskever told The Verge in an interview published this week. "GPT-4 is not easy to develop...there are many many companies who want to do the same thing, so from a competitive side, you can see this as a maturation of the field." & Chat GPT is an attempt to make AI into a conversation between humans and the technology via the computer. People ask it questions and get text message-style answers. Regarding Chat GPT, Marks dismissed it as unreliable: "They don't tell the truth. In fact, Chat GPT, when you log on to it, says... 'Don't trust the facts that we're telling you.'" (The full chat GPT warning reads: "While we have safeguards in place, the system may occasionally generate incorrect or misleading information and produce offensive or biased content. It is not intended to give advice.")AI EXPERTS WEIGH DANGERS, BENEFITS OF CHATGPT ON HUMANS, JOBS AND INFORMATION: 'DYSTOPIAN WORLD'Marks said of Chat GPT-style software: "They aren't creative. They don't understand. They have a terrible sense of humor.” \\
\addlinespace
\bottomrule 
\end{tabularx}
\end{table*}
\onecolumn

\section{Tuning of $\lambda$}
\label{app:lambda}

We conduct experiments by varying $\lambda$ across a broad range. As shown in Table~\ref{tab:lambda_a} and Table~\ref{tab:lambda_b}, $\lambda=0.5$ provides a balanced trade-off, achieving strong performance on relevance metrics (P, R, F1) while preserving diversity (D, I, C). 


\begin{table*}[h]
\centering
\begin{tabular}{@{}c cccccc cccccc@{}}
\toprule
\multirow{3}{*}{\boldmath$\lambda$} 
& \multicolumn{3}{c}{\textbf{Relevancy}} & \multicolumn{3}{c}{\textbf{Diversity}} 
& \multicolumn{3}{c}{\textbf{Relevancy}} & \multicolumn{3}{c}{\textbf{Diversity}} \\
\cmidrule(lr){2-4} \cmidrule(lr){5-7} \cmidrule(lr){8-10} \cmidrule(lr){11-13}
 & \textbf{P} & \textbf{R} & \textbf{F1} & \textbf{D} & \textbf{I} & \textbf{C} 
 & \textbf{P} & \textbf{R} & \textbf{F1} & \textbf{D} & \textbf{I} & \textbf{C} \\ 
 \cmidrule(lr){2-7}\cmidrule(lr){8-13}
 & \multicolumn{6}{c}{\textbf{top 5}} & \multicolumn{6}{c}{\textbf{top 10}} \\ 
 \midrule
1/8 & 90.5 & 15.7 & 26.7 & \textbf{31.4} & \textbf{74.9} & \textbf{64.0} & 86.9 & 30.2 & 44.8 & \textbf{32.5} & \textbf{58.0} & \textbf{78.8} \\
1/4 & 91.5 & 16.0 & 27.2 & 30.7 & 74.3 & 63.3 & 89.2 & 31.3 & 46.3 & 31.3 & 56.1 & 76.9 \\
1/2 & 92.4 & 16.2 & 27.5 & 29.8 & 73.6 & 62.2 & 90.3 & 31.6 & 46.8 & 30.2 & 54.4 & 74.5 \\
1   & 92.8 & 16.2 & 27.6 & 29.0 & 72.7 & 61.8 & 91.3 & 32.0 & 47.4 & 29.5 & 52.3 & 72.4 \\
2   & 93.8 & 16.5 & 28.1 & 27.8 & 69.7 & 58.8 & 92.4 & 32.3 & 47.9 & 28.6 & 49.6 & 69.1 \\
4   & 94.8 & 16.7 & 28.4 & 26.9 & 68.7 & 57.2 & 93.2 & \textbf{32.6} & 48.3 & 27.9 & 48.5 & 66.7 \\
8   & \textbf{95.5} & \textbf{16.9} & \textbf{28.7} & 25.3 & 66.5 & 54.6 & \textbf{93.4} & \textbf{32.6} & \textbf{48.4} & 27.0 & 46.0 & 63.6 \\ 
 \midrule
 & \multicolumn{6}{c }{\textbf{top 20}} & \multicolumn{6}{c}{\textbf{top 50}} \\ \midrule
1/8 & 81.3 & 54.3 & 65.1 & \textbf{36.1} & \textbf{43.2} & \textbf{87.3} & 53.8 & 77.6 & 63.5 & \textbf{47.9} & \textbf{35.3} & \textbf{93.4} \\
1/4 & 82.8 & 55.1 & 66.2 & 35.4 & 40.9 & 86.4 & 53.9 & 77.7 & 63.7 & 47.7 & 34.0 & 93.2 \\
1/2 & 84.1 & 55.8 & 67.1 & 34.9 & 38.8 & 84.7 & 54.1 & 78.0 & 63.9 & 47.6 & 32.2 & 92.7 \\
1   & 84.8 & 56.2 & 67.6 & 34.5 & 36.7 & 82.6 & 54.3 & 78.3 & 64.1 & 47.4 & 30.8 & 92.5 \\
2   & 85.2 & 56.3 & 67.8 & 34.2 & 35.2 & 81.1 & 54.4 & 78.4 & \textbf{64.3} & 47.3 & 29.5 & 92.2 \\
4   & 85.6 & 56.4 & 68.0 & 34.0 & 33.7 & 79.3 & 54.4 & 78.5 & \textbf{64.3} & 47.3 & 28.5 & 91.7 \\
8   & \textbf{85.8} & \textbf{56.6} & \textbf{68.2} & 33.8 & 33.0 & 78.4 & \textbf{54.5} & \textbf{78.5} & \textbf{64.3} & 47.2 & 27.9 & 91.6 \\ 
\bottomrule
\end{tabular}
\caption{Performance on \textbf{LocalNews} benchmark across different $\lambda$ values. Best values in each column are boldfaced.}
\label{tab:lambda_a}
\end{table*}

\begin{table*}[h]
\centering
\begin{tabular}{@{}c cccccc cccccc@{}}
\toprule
\multirow{3}{*}{\boldmath$\lambda$} 
& \multicolumn{3}{c}{\textbf{Relevancy}} & \multicolumn{3}{c}{\textbf{Diversity}} 
& \multicolumn{3}{c}{\textbf{Relevancy}} & \multicolumn{3}{c}{\textbf{Diversity}} \\
\cmidrule(lr){2-4} \cmidrule(lr){5-7} \cmidrule(lr){8-10} \cmidrule(lr){11-13}
 & \textbf{P} & \textbf{R} & \textbf{F1} & \textbf{D} & \textbf{I} & \textbf{C} 
 & \textbf{P} & \textbf{R} & \textbf{F1} & \textbf{D} & \textbf{I} & \textbf{C} \\ 
 \cmidrule(lr){2-7}\cmidrule(lr){8-13}
 & \multicolumn{6}{c}{\textbf{top 5}} & \multicolumn{6}{c}{\textbf{top 10}} \\ 
 \midrule
1/8 & 92.5 & \textbf{10.5} & \textbf{18.9} & 33.2 & 68.6 & 41.2 & 89.3 & 20.2 & 33.0 & \textbf{36.0} & 54.2 & 59.6 \\
1/4 & 92.5 & \textbf{10.5} & \textbf{18.9} & 33.0 & 68.4 & 40.9 & 89.6 & 20.3 & 33.1 & 35.4 & 53.8 & 58.7 \\
1/2 & 91.6 & 10.4 & 18.7 & \textbf{33.8} & \textbf{74.4} & \textbf{44.6} & 89.3 & 20.2 & 33.0 & 34.9 & \textbf{57.4} & \textbf{60.8} \\
1   & 92.1 & \textbf{10.5} & 18.8 & 32.5 & 67.8 & 39.9 & 90.5 & 20.5 & 33.5 & 34.2 & 52.3 & 56.8 \\
2   & 92.1 & \textbf{10.5} & 18.8 & 31.7 & 66.9 & 39.5 & 91.2 & 20.7 & 33.7 & 33.2 & 50.0 & 55.3 \\
4   & 92.0 & \textbf{10.5} & 18.8 & 31.0 & 64.8 & 38.0 & 91.8 & 20.8 & 33.9 & 32.1 & 48.3 & 53.2 \\
8   & \textbf{92.7} & \textbf{10.5} & \textbf{18.9} & 30.3 & 64.2 & 36.9 & \textbf{92.0} & \textbf{20.9} & \textbf{34.0} & 31.0 & 46.3 & 50.9 \\ 
 \midrule
 & \multicolumn{6}{c }{\textbf{top 20}} & \multicolumn{6}{c}{\textbf{top 50}} \\ \midrule
1/8 & 85.3 & 38.5 & 53.1 & \textbf{38.4} & \textbf{38.8} & \textbf{76.5} & 69.1 & 73.9 & 71.4 & \textbf{47.7} & \textbf{27.1} & 93.1 \\
1/4 & 86.8 & 39.2 & 54.0 & 37.2 & 37.8 & 75.0 & 70.0 & 74.8 & 72.3 & 47.5 & 26.4 & \textbf{93.2} \\
1/2 & 88.3 & 39.9 & 55.0 & 35.7 & 38.3 & 74.0 & 71.2 & 76.1 & 73.6 & 47.1 & 26.3 & 93.1 \\
1   & 88.5 & 40.0 & 55.1 & 35.5 & 35.4 & 72.3 & 71.3 & 76.1 & 73.6 & 47.0 & 25.0 & 92.6 \\
2   & 89.6 & 40.4 & 55.7 & 34.9 & 33.9 & 70.4 & 71.8 & 76.7 & 74.2 & 46.9 & 24.3 & 92.4 \\
4   & 90.1 & 40.7 & 56.1 & 34.4 & 32.9 & 68.6 & 71.9 & 76.8 & \textbf{74.3} & 46.8 & 23.8 & 92.3 \\
8   & \textbf{90.4} & \textbf{40.8} & \textbf{56.3} & 34.2 & 32.1 & 67.4 & \textbf{72.0} & \textbf{76.9} & \textbf{74.3} & 46.8 & 23.4 & 92.2 \\ 
\bottomrule
\end{tabular}
\caption{Performance on \textbf{DSGlobal} benchmark across different $\lambda$ values.}
\label{tab:lambda_b}
\end{table*}

\section{Ablation Study}

\subsection{Full Results}
\label{sec:appendix_ablation_result}

We present the complete ablation results for \textbf{NEWSCOPE (GreedyPlus)} on \textbf{LocalNews} (Table~\ref{tab:ablation_local}) and \textbf{DSGlobal} (Table~\ref{tab:ablation_global}). \textbf{Removing the diversity term} improves F1 slightly but leads to sharp drops in D, I, and C, indicating increased redundancy and weaker perspective coverage. \textbf{Removing the relevance term} boosts diversity (with top scores in D, I, and C) but significantly lowers F1, suggesting inclusion of less relevant content. These findings reinforce that both relevance and diversity are essential. 

\begin{table*}[th]
\centering
\small
\begin{tabular}{@{}c cccccc cccccc@{}}
\toprule
\multirow{2}{*}{\textbf{Model}} & \textbf{P} & \textbf{R} & \textbf{F1} & \textbf{D} & \textbf{I} & \textbf{C} & \textbf{P} & \textbf{R} & \textbf{F1} & \textbf{D} & \textbf{I} & \textbf{C} \\ \cmidrule(lr){2-7}\cmidrule(lr){8-13}
 & \multicolumn{6}{c}{\textbf{top 5}} & \multicolumn{6}{c}{\textbf{top 10}} \\ \midrule
\textbf{NEWSCOPE (GreedyPlus)} & 92.4 & 16.2 & 27.5 & 29.8 & 73.6 & 62.2 & 90.3 & 31.6 & 46.8 & 30.2 & 54.4 & 74.5 \\
\textbf{w/o diversity term} & \textbf{97.3} & \textbf{17.2} & \textbf{29.3} & 17.7 & 45.1 & 35.8 & \textbf{94.6} & \textbf{32.9} & \textbf{48.8} & 24.1 & 36.1 & 52.8 \\
\textbf{w/o relevance term} & 87.6 & 14.7 & 25.2 & \textbf{34.4} & \textbf{82.3} & \textbf{64.9} & 68.3 & 20.9 & 32.0 & \textbf{42.9} & \textbf{76.0} & \textbf{80.5} \\ \midrule
 & \multicolumn{6}{c}{\textbf{top 20}} & \multicolumn{6}{c}{\textbf{top 50}} \\ \midrule
\textbf{NEWSCOPE (GreedyPlus)} & 84.1 & 55.8 & 67.1 & 34.9 & 38.8 & 84.7 & 54.1 & 78.0 & 63.9 & 47.6 & 32.2 & 92.7 \\
\textbf{w/o diversity term} & \textbf{86.1} & \textbf{56.6} & \textbf{68.3} & 33.3 & 30.1 & 75.0 & \textbf{54.5} & \textbf{78.6} & \textbf{64.4} & 47.2 & 26.7 & 90.9 \\
\textbf{w/o relevance term} & 43.2 & 23.8 & 30.7 & \textbf{51.0} & \textbf{72.2} & \textbf{87.4} & 48.5 & 70.3 & 57.4 & \textbf{49.2} & \textbf{42.7} & \textbf{93.1} \\
\bottomrule
\end{tabular}
\caption{Ablation study of NEWSCOPE (GreedyPlus) on the \textbf{LocalNews} benchmark.}
\label{tab:ablation_local}
\end{table*}

\begin{table*}[th]
\centering
\small
\begin{tabular}{@{}c cccccc cccccc@{}}
\toprule
\multirow{2}{*}{\textbf{Model}} & \textbf{P} & \textbf{R} & \textbf{F1} & \textbf{D} & \textbf{I} & \textbf{C} & \textbf{P} & \textbf{R} & \textbf{F1} & \textbf{D} & \textbf{I} & \textbf{C} \\ \cmidrule(lr){2-7}\cmidrule(lr){8-13}
 & \multicolumn{6}{c}{\textbf{top 5}} & \multicolumn{6}{c}{\textbf{top 10}} \\ \midrule
\textbf{NEWSCOPE (GreedyPlus)} & 91.6 & 10.4 & 18.7 & 33.8 & 74.4 & 44.6 & 89.3 & 20.2 & 33.0 & 34.9 & 57.4 & 60.8 \\
\textbf{w/o diversity term} & \textbf{95.8} & \textbf{10.8} & \textbf{19.4} & 22.2 & 46.6 & 24.3 & \textbf{93.6} & \textbf{21.2} & \textbf{34.6} & 26.9 & 37.2 & 39.9 \\
\textbf{w/o relevance term} & 90.7 & 10.3 & 18.5 & \textbf{35.1} & \textbf{76.9} & \textbf{45.1} & 83.9 & 18.5 & 30.4 & \textbf{41.5} & \textbf{66.5} & \textbf{63.4} \\ \midrule
 & \multicolumn{6}{c}{\textbf{top 20}} & \multicolumn{6}{c}{\textbf{top 50}} \\ \midrule
\textbf{NEWSCOPE (GreedyPlus)} & 88.3 & 39.9 & 55.0 & 35.7 & 38.3 & 74.0 & 71.2 & 76.1 & 73.6 & 47.1 & 26.3 & \textbf{93.1} \\
\textbf{w/o diversity term} & \textbf{90.5} & \textbf{40.9} & \textbf{56.3} & 33.5 & 29.8 & 64.1 & \textbf{72.1} & \textbf{77.0} & \textbf{74.5} & 46.7 & 22.8 & 91.7 \\
\textbf{w/o relevance term} & 60.2 & 25.2 & 35.5 & \textbf{51.3} & \textbf{58.4} & \textbf{75.1} & 
57.0 & 61.4 & 59.1 & \textbf{52.3} & \textbf{36.0} & 90.3 \\
\bottomrule
\end{tabular}
\caption{Ablation study of NEWSCOPE (GreedyPlus) on the \textbf{DSGlobal} benchmark.}
\label{tab:ablation_global}
\end{table*}

\subsection{Qualitative Analysis for Ablation Study}
\label{sec:appendix_ablation}

To further support the findings in Section 6.3, we provide a qualitative comparison of top-20 retrieved results for each ablation variant under the DSGlobal benchmark.

\paragraph{NEWSCOPE(GreedyPlus).} Retrieves content spanning multiple dimensions, including Altman’s interviews, global reactions, adoption benchmarks, and criticism, offering comprehensive coverage.

\paragraph{w/o Diversity Term.} This reduces to a relevance-only model. The retrieved paragraphs are repetitive, with most centered around similar quotes from Altman. This improves precision but fails to uncover distinct viewpoints.

\paragraph{w/o Relevance Term.} This variant retrieves many semantically distinct paragraphs, including out-of-context or weakly related content (e.g., speculative commentary, minor international opinions). While diverse, it lacks cohesion and topical alignment.

These patterns are reflected quantitatively in the main paper and qualitatively here, reinforcing that both scoring components are necessary to balance informativeness and perspective diversity.

\begin{table*}[htbp]
\caption{Example: Top-20 Paragraphs Retrieved by Three Strategies}
\small
\newcolumntype{N}{>{\centering\arraybackslash}p{0.4cm}} 
\newcolumntype{M}{>{\arraybackslash}p{4cm}} 
\begin{tabularx}{\linewidth}{@{} N M >{\RaggedRight\arraybackslash}X >{\RaggedRight\arraybackslash}X @{}}

\toprule 

\multicolumn{1}{c}{\textbf{Rank}} & \multicolumn{1}{c}{\textbf{w/o diversity term}} & \multicolumn{1}{c}{\textbf{NEWSCOPE (GreedyPlus)}} & \multicolumn{1}{c}{\textbf{w/o relevance term}} \\
\midrule
\multicolumn{1}{c}{\textbf{1}} & Archaeologists discovered the oldest pearling town in the Persian Gulf on Siniyah Island in Umm al-Quwain, dating back to the late 6th century. & \textbf{Ranked 1 by sim\_score.} Archaeologists discovered the oldest pearling town in the Persian Gulf on Siniyah Island in Umm al-Quwain, dating back to the late 6th century. & \textbf{Ranked 1 by sim\_score.} Archaeologists discovered the oldest pearling town in the Persian Gulf on Siniyah Island in Umm al-Quwain, dating back to the late 6th century. \\
\addlinespace 
\multicolumn{1}{c}{\textbf{2}}  & Timothy Power claimed that this is the oldest example of that kind of very specifically Khaleeji pearling town. & \textbf{Ranked 21 by sim\_score.} Houses in the pearling town are densely packed. Umm al-Quwain plans to build a visitor's center at the site. & \textbf{Ranked 21 by sim\_score.} Houses in the pearling town are densely packed. Umm al-Quwain plans to build a visitor's center at the site. \\
\addlinespace 
\multicolumn{1}{c}{\textbf{3}}  & \textbf{same as above} & \textbf{Ranked 14 by sim\_score.} The pearling town on Siniyah Island, near an ancient Christian monastery, spans 12 hectares and shows evidence of year-round habitation and social stratification. & \textbf{Ranked 14 by sim\_score.} The pearling town on Siniyah Island, near an ancient Christian monastery, spans 12 hectares and shows evidence of year-round habitation and social stratification. \\
\addlinespace 
\multicolumn{1}{c}{\textbf{4}}  & \textbf{same as above} & \textbf{Ranked 11 by sim\_score.} Archaeologists discovered a 1,300-year-old pearling town on Siniya Island in Umm Al Quwain, marking the oldest such site in the Arabian Gulf with year-round habitation. & \textbf{Ranked 11 by sim\_score.} Archaeologists discovered a 1,300-year-old pearling town on Siniya Island in Umm Al Quwain, marking the oldest such site in the Arabian Gulf with year-round habitation. \\
\addlinespace 
\multicolumn{1}{c}{\textbf{5}}  & \textbf{same as above} & \textbf{Ranked 2 by sim\_score.} Timothy Power claimed that this is the oldest example of that kind of very specifically Khaleeji pearling town. & \textbf{Ranked 2 by sim\_score.} Timothy Power claimed that this is the oldest example of that kind of very specifically Khaleeji pearling town. \\
\addlinespace 
\multicolumn{1}{c}{\textbf{6}}  & \textbf{same as above} & \textbf{Ranked 12 by sim\_score.} The oldest known Khaleeji pearling town, found on Siniyah Island in Umm al-Quwain, predates Dubai and is near an ancient Christian monastery. & \textbf{Ranked 12 by sim\_score.} The oldest known Khaleeji pearling town, found on Siniyah Island in Umm al-Quwain, predates Dubai and is near an ancient Christian monastery. \\
\addlinespace 
\multicolumn{1}{c}{\textbf{7}}  & \textbf{same as above} & \textbf{Ranked 31 by sim\_score.} The collapse of pearling after WWI mirrors the UAE's challenge of transitioning from fossil fuels to a carbon-neutral future, as highlighted by discoveries of oyster shell remains on Siniya Island. & \textbf{Ranked 40 by sim\_score.} Multiple organizations excavated the ancient pearling town, with plans for a visitor's center amid Umm al-Quwain's efforts to balance development, including a \$675 million project, with lessons from its historical sites. \\
\addlinespace 
\multicolumn{1}{c}{\textbf{8}}  & \textbf{same as above} & \textbf{Ranked 28 by sim\_score.} The pearling town on Siniya Island, which protects the Khor Al Beida marshlands in Umm Al Quwain, features year-round habitation and artifacts from the pearling industry, alongside a 1,400-year-old Christian monastery. & \textbf{Ranked 28 by sim\_score.} The pearling town on Siniya Island, which protects the Khor Al Beida marshlands in Umm Al Quwain, features year-round habitation and artifacts from the pearling industry, alongside a 1,400-year-old Christian monastery. \\
\addlinespace 
\multicolumn{1}{c}{\textbf{9}}  & Archaeologists discovered the oldest pearling town in the Persian Gulf on Siniyah Island in Umm al-Quwain, UAE, dating back to the late 6th century. & \textbf{Ranked 30 by sim\_score.} The pearling town, spanning 12 hectares south of an ancient monastery, features diverse homes indicating social stratification and year-round habitation, with artifacts like pearls and diving weights found inside. & \textbf{Ranked 10 by sim\_score.} The discovery of a 12-hectare pearling town on Al Siniya Island, dating to the pre-Islamic era, highlights Umm Al Quwain's thriving pearl trade and historical significance. \\
\addlinespace 
\multicolumn{1}{c}{\textbf{10}}  & The discovery of a 12-hectare pearling town on Al Siniya Island, dating to the pre-Islamic era, highlights Umm Al Quwain's thriving pearl trade and historical significance. & \textbf{Ranked 40 by sim\_score.}Multiple organizations excavated the ancient pearling town, with plans for a visitor's center amid Umm al-Quwain's efforts to balance development, including a \$675 million project, with lessons from its historical sites. & \textbf{Ranked 31 by sim\_score.} The collapse of pearling after WWI mirrors the UAE's challenge of transitioning from fossil fuels to a carbon-neutral future, as highlighted by discoveries of oyster shell remains on Siniya Island. \\
\addlinespace 
\bottomrule
\multicolumn{4}{r}{\textit{Table continued in next page.}} \\*
\end{tabularx}
\end{table*}
\newpage

\begin{table*}[htbp]
\small
\newcolumntype{N}{>{\centering\arraybackslash}p{0.4cm}} 
\newcolumntype{M}{>{\arraybackslash}p{4cm}} 
\begin{tabularx}{\linewidth}{@{} N M >{\RaggedRight\arraybackslash}X >{\RaggedRight\arraybackslash}X @{}}

\toprule 

\multicolumn{1}{c}{\textbf{Rank}} & \multicolumn{1}{c}{\textbf{w/o diversity term}} & \multicolumn{1}{c}{\textbf{NEWSCOPE (GreedyPlus)}} & \multicolumn{1}{c}{\textbf{w/o relevance term}} \\
\midrule
\multicolumn{1}{c}{\textbf{11}}  & Archaeologists discovered a 1,300-year-old pearling town on Siniya Island in Umm Al Quwain, marking the oldest such site in the Arabian Gulf with year-round habitation. & \textbf{Ranked 10 by sim\_score.} The discovery of a 12-hectare pearling town on Al Siniya Island, dating to the pre-Islamic era, highlights Umm Al Quwain's thriving pearl trade and historical significance. & \textbf{Ranked 44 by sim\_score.} Umm al-Quwain plans a \$675 million real estate development including a bridge to Siniyah Island, hoping to boost the economy while learning from ancient pearling sites. \\
\addlinespace 
\multicolumn{1}{c}{\textbf{12}}  & The oldest known Khaleeji pearling town, found on Siniyah Island in Umm al-Quwain, predates Dubai and is near an ancient Christian monastery. & \textbf{Ranked 2 by sim\_score.} & \textbf{irrelevant} \\
\addlinespace 
\multicolumn{1}{c}{\textbf{13}}  & \textbf{same as above} & \textbf{Ranked 2 by sim\_score.} & \textbf{irrelevant} \\
\addlinespace 
\multicolumn{1}{c}{\textbf{14}}  & The pearling town on Siniyah Island, near an ancient Christian monastery, spans 12 hectares and shows evidence of year-round habitation and social stratification. & \textbf{Ranked 2 by sim\_score.} & \textbf{irrelevant} \\
\addlinespace 
\multicolumn{1}{c}{\textbf{15}} & \textbf{same as above} & \textbf{Ranked 2 by sim\_score.} & \textbf{irrelevant} \\
\addlinespace 
\multicolumn{1}{c}{\textbf{16}}  & \textbf{same as above} & \textbf{Ranked 2 by sim\_score.} & \textbf{irrelevant} \\
\addlinespace 
\multicolumn{1}{c}{\textbf{17}}  & \textbf{same as above} & \textbf{Ranked 12 by sim\_score.} & \textbf{irrelevant} \\
\addlinespace 
\multicolumn{1}{c}{\textbf{18}}  & \textbf{same as above} & \textbf{Ranked 14 by sim\_score.} & \textbf{irrelevant} \\
\addlinespace 
\multicolumn{1}{c}{\textbf{19}}  & \textbf{same as above} & \textbf{Ranked 14 by sim\_score.} & \textbf{irrelevant} \\
\addlinespace 
\multicolumn{1}{c}{\textbf{20}}  & \textbf{same as above} & \textbf{Ranked 14 by sim\_score.} & \textbf{irrelevant} \\
\addlinespace 
\bottomrule
\end{tabularx}
\end{table*}

\end{document}